\documentclass[10pt,letterpaper]{article}
\usepackage[T1]{fontenc}
\usepackage{times}
\usepackage[letterpaper,left=0.72in,right=0.72in,top=0.72in,bottom=0.80in,footskip=0.34in]{geometry}
\usepackage{amsmath,amssymb,amsthm}
\usepackage{graphicx,xcolor,booktabs,multirow,needspace}
\usepackage{multicol}
\usepackage[font=small,labelfont=bf,labelsep=period]{caption}
\usepackage[numbers,sort&compress]{natbib}
\usepackage[hyphens]{url}
\usepackage[breaklinks,colorlinks,bookmarksnumbered]{hyperref}
\definecolor{linkink}{HTML}{234F65}
\hypersetup{
  allcolors=linkink,
  pdftitle={PerturbJEPA: Joint-Embedding Predictive Learning of Cellular Perturbation Responses},
  pdfauthor={Jiangshuan Pang; Youzhe He; Yiting Deng; Wangyang Tang; Huinan Xu; Qi Lu; Shiping Liu; Zhenkun Zhuang},
  pdfsubject={Preprint on intervention retrieval from cellular perturbation responses}
}
\newtheorem{proposition}{Proposition}
\newcommand{\etal}{\emph{et al.}}

\makeatletter
\renewcommand{\section}{\@startsection{section}{1}{\z@}
  {-3ex plus -0.8ex minus -0.2ex}{1.6ex plus 0.2ex}{\normalfont\Large\bfseries\raggedright}}
\renewcommand{\subsection}{\@startsection{subsection}{2}{\z@}
  {-2.7ex plus -0.7ex minus -0.2ex}{1.2ex plus 0.2ex}{\normalfont\large\bfseries\raggedright}}
\makeatother

\begin{document}
\onecolumn
\thispagestyle{plain}
\pdfbookmark[1]{Abstract}{abstract}
\begingroup
  \centering
  \setlength{\parindent}{0pt}
  \setlength{\parskip}{0pt}
  \makebox[0.92\linewidth]{%
    \raisebox{-0.5\height}{\includegraphics[height=11mm]{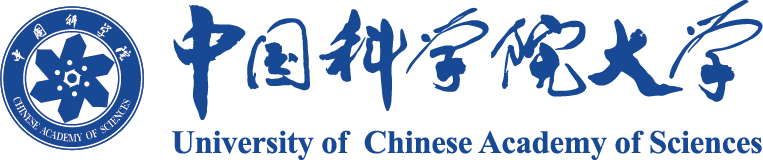}}\hfill
    \raisebox{-0.5\height}{\includegraphics[height=11mm]{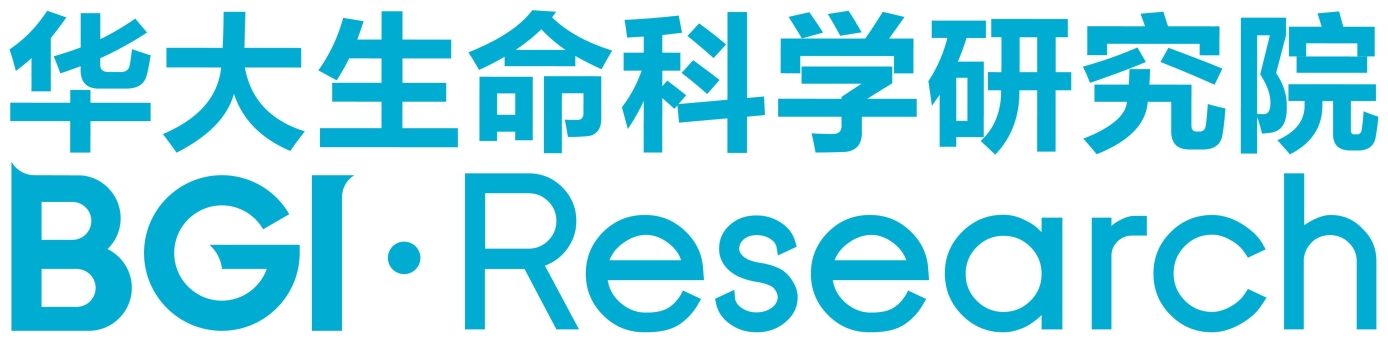}}\hfill
    \raisebox{-0.5\height}{\includegraphics[height=18mm]{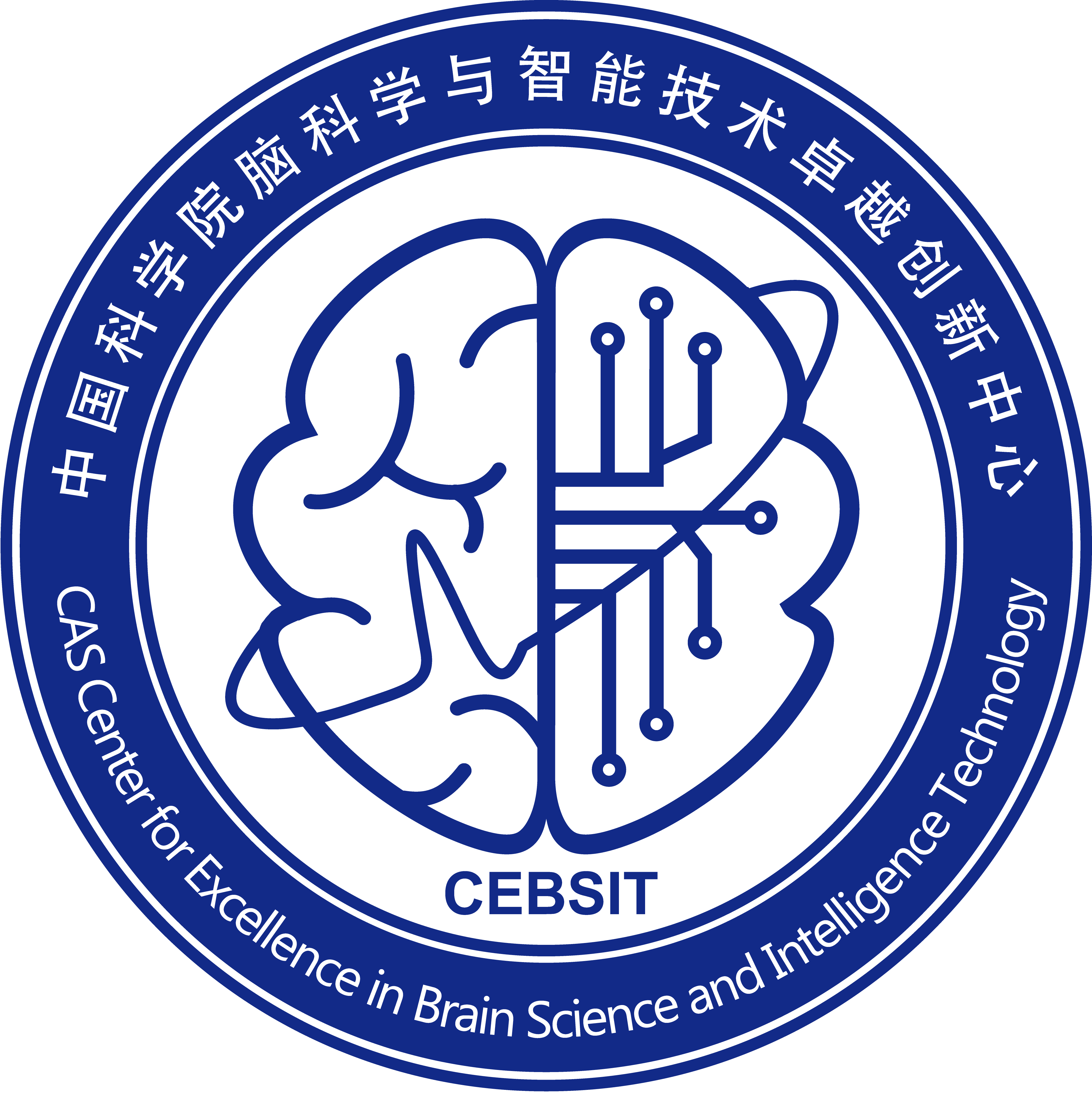}}}\par
  \vspace{4mm}
  {\color[gray]{.72}\hrule height .4pt}
  \vspace{9mm}
  {\fontsize{21}{25}\selectfont\bfseries
    PerturbJEPA: Joint-Embedding Predictive Learning\\
    of Cellular Perturbation Responses\par}
  \vspace{6mm}
  {\fontsize{12}{16}\selectfont
    Jiangshuan Pang\textsuperscript{1,2}, Youzhe He\textsuperscript{2},
    Yiting Deng\textsuperscript{1,2}, Wangyang Tang\textsuperscript{2}\\
    Huinan Xu\textsuperscript{2}, Qi Lu\textsuperscript{3},
    Shiping Liu\textsuperscript{2,*}, Zhenkun Zhuang\textsuperscript{2,*}\par}
  \vspace{4mm}
  {\fontsize{10}{13}\selectfont
    \textsuperscript{1}School of Artificial Intelligence, University of Chinese Academy of Sciences, Beijing 100049, China\\
    \textsuperscript{2}Key Laboratory of Brain Cell Mapping of Zhejiang Province, BGI Research, Hangzhou 310030, China.\\
    \textsuperscript{3}Center for Excellence in Brain Science and Intelligence Technology
    (Institute of Neuroscience),\\Chinese Academy of Sciences\par}
  \vspace{2mm}
  {\small\urlstyle{same}\textsuperscript{*}Corresponding authors:
    \href{mailto:liushiping@genomics.cn}{\nolinkurl{liushiping@genomics.cn}},
    \href{mailto:zhuangzhenkun@genomics.cn}{\nolinkurl{zhuangzhenkun@genomics.cn}}\par}
  \vspace{9mm}
  {\Large\bfseries Abstract\par}
  \vspace{3mm}
  \begin{minipage}{0.82\linewidth}
    \setlength{\parindent}{0pt}%
    \fontsize{11.5}{16.4}\selectfont Identifying the drug and dose that produced a cellular response requires preserving treatment effects within each cellular background.
A joint-embedding predictor can match its targets while ignoring the intervention if its encoder preserves differences across backgrounds but erases differences between treated and control populations within each one.
We call this failure \emph{interventional contrast collapse}.
\mbox{PerturbJEPA} encodes the difference in mean gene expression between treated and control populations as an effect vector.
It aligns the displacement between their latent centroids with this vector and regularizes the vectors to preserve variation across population pairs.
We show that the measured displacements cannot all coincide within a batch when effect-vector variation exceeds the root-mean-square alignment error.
Trained with separate drug and dose ranking objectives, the model predicts candidate responses from a common control population and matches them to the observed response without cell pairing or expression reconstruction.
We evaluate 3,592 Tahoe-100M queries on held-out combinations of known interventions and cellular backgrounds, each with over 700 drug--dose candidates in a fixed, background-specific catalog.
\mbox{PerturbJEPA} achieves 91.46\% macro joint drug--dose Recall@1, exceeding STATE by 3.40 percentage points (95\% drug-cluster bootstrap interval of 1.72 to 5.07 points).
Controlled diagnostics show that treatment--control differences can shrink even when representations remain globally dispersed.
With the drug ranking objective fixed, adding effect-vector regularization to alignment raises joint Recall@1 from 18.57\% to 87.47\%.
\par
    \vspace{3mm}
    {\small\urlstyle{same}\textbf{Code:} \url{https://github.com/BGI-Hangzhou-OmicsAI/PerturbJEPA}\par}
  \end{minipage}\par
\endgroup
\twocolumn

\section{Introduction}
\label{sec:introduction}

Recovering the drug and dose responsible for an observed cellular response requires distinguishing intervention effects within the same cellular background.
We study \emph{source-intervention retrieval}: a forward model predicts each candidate response from a common control population and retrieves the closest match to an observed target from a catalog containing the true intervention.

\suppressfloats[t]
\begin{figure}[t]
\centering
\includegraphics[width=\columnwidth]{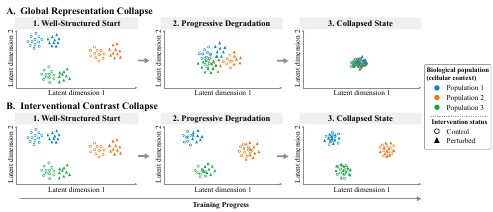}
\caption{\textbf{Global collapse and interventional contrast collapse are different failure modes.}
Colors denote cellular backgrounds; circles and triangles denote control and treated cells.
(A) Global collapse erases both background variation and intervention contrasts.
(B) ICC removes treatment--control separation within each background while retaining global dispersion.
These are conceptual trajectories; Fig.~\ref{fig:mechanism} reports measured training dynamics.}
\label{fig:collapse-schematic}
\end{figure}

\begin{figure*}[t]
\centering
\includegraphics[width=\textwidth]{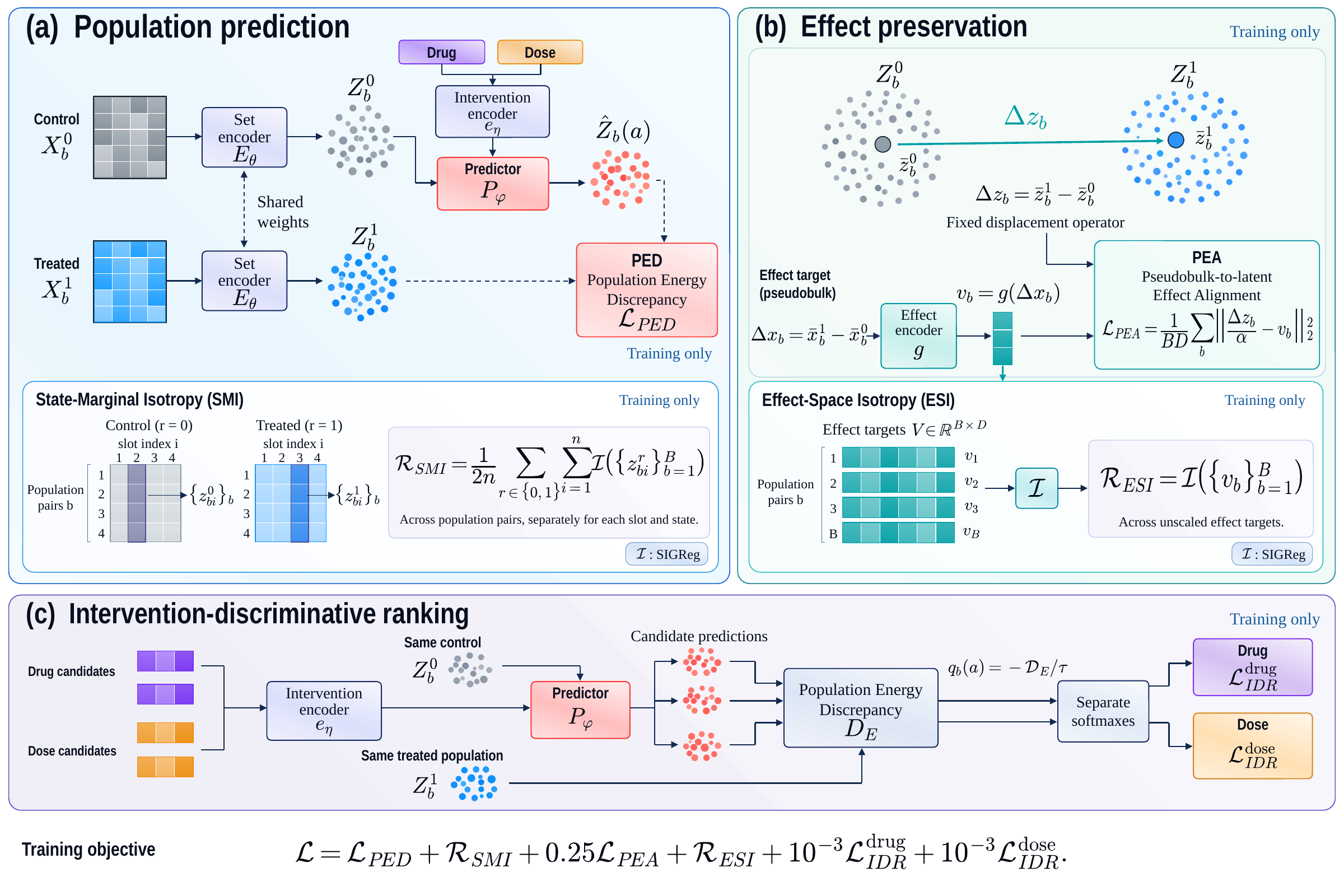}
\caption{\textbf{PerturbJEPA learns cell population responses for source-intervention retrieval.}
\textbf{(a) Population prediction.} A shared set encoder embeds unpaired control and treated cell populations. An intervention-conditioned predictor minimizes population energy discrepancy (PED); state-marginal isotropy (SMI) regularizes each slot across population pairs, separately for each state.
\textbf{(b) Effect preservation.} Pseudobulk-to-latent effect alignment (PEA) aligns the scaled, fixed centroid displacement $\Delta z_b/\alpha$ with the expression-contrast embedding $v_b=g(\Delta x_b)$. Effect-space isotropy (ESI) regularizes these unscaled effect targets.
\textbf{(c) Intervention-discriminative ranking (IDR).} Candidate predictions share the same control and are compared with the same treated population, using separate drug and dose softmax losses.
Point clouds and matrices are schematic. The effect branch and loss terms are training-only; retrieval scores candidates by PED.}
\label{fig:architecture}
\end{figure*}

This extends transcriptional signature matching~\citep{lamb2006connectivity,subramanian2017l1000} by conditioning predictions on the observed control cells.
Destructive single-cell measurements preclude tracking individual cells before and after treatment, requiring prediction between unpaired populations.
Joint-embedding predictive architectures (JEPAs) learn to predict target representations without reconstructing observations~\citep{assran2025vjepa2}.
Yet accurate prediction can become uninformative for retrieval if the representation discards treatment effects while preserving larger background variation.

\textbf{Global representation diversity is insufficient.}
Marginal regularizers discourage a constant representation by maintaining variation across examples~\citep{balestriero2025lejepa,maes2026leworldmodel}.
Yet an encoder can preserve differences between cellular backgrounds while assigning the same representation distribution to control and treated populations within each background.
We call this failure \emph{interventional contrast collapse} (ICC; Fig.~\ref{fig:collapse-schematic}).
Under ICC, returning the control representation can match the target distribution. If the predictor does so for two doses of the same drug, their scores tie. The cellular background is identical for both candidates.

\textbf{Effect preservation in retrieval geometry.}
Although each cell is measured only once, the treated and control means provide a measurable expression contrast without cell pairing.
\emph{Pseudobulk-to-latent effect alignment} (PEA) uses this contrast to supervise retrieval geometry (Fig.~\ref{fig:architecture}(b)). It aligns the scaled displacement between the two latent centroids with a learned embedding of the expression contrast.
An auxiliary head could amplify a small displacement to predict the same target, even if the centroids remain close. Direct alignment ties the target to the displacement itself.
Regularizing the effect target is essential: alignment alone allows the target and displacement to shrink together.

A finite-batch bound relates effect-target variation and alignment error to measured displacement variation across population pairs.
Controlled diagnostics assess treatment--control contrast during training, and separate drug and dose ranking losses train candidate discrimination.
Population energy discrepancy (PED) additionally matches distributional structure and scores candidates at inference, when the effect branch is removed.

We evaluate on $3{,}592$ Tahoe-100M queries~\citep{zhang2026tahoe} with held-out combinations of known interventions and cellular backgrounds, each ranking over $700$ drug--dose candidates.
PerturbJEPA reaches 91.46\% macro joint Recall@1, exceeding STATE by 3.40 percentage points.
With the drug-ranking objective unchanged, adding effect-target regularization to alignment raises joint Recall@1 from 18.57\% to 87.47\%.
\mbox{We make three contributions.}
\begin{itemize}
\item We characterize ICC as a failure of conditional response retention despite global diversity, and diagnose contrast contraction under controlled training.
\item We introduce a fixed population-displacement constraint with regularized effect targets and establish a finite-batch bound on measured displacement variation.
\item We show that retaining intervention contrasts and learning candidate orderings provide complementary retrieval gains on held-out combinations of known factors.
\end{itemize}

\section{Related work}
\label{sec:related-work}

\paragraph{Predictive representations and selective feature loss.}
V-JEPA~2 learns predictive representations without observation reconstruction and supports action-conditioned prediction and planning~\cite{assran2025vjepa2}.
LeJEPA~\cite{balestriero2025lejepa} introduces SIGReg to encourage an isotropic Gaussian embedding distribution, while LeWorldModel~\cite{maes2026leworldmodel} combines distributional regularization with end-to-end latent dynamics learning.
These objectives address global representation collapse, but preserving marginal diversity does not ensure that every relevant factor survives.
Sobal \etal~\cite{sobal2022slow} show that predictive embeddings can favor persistent backgrounds over changing features.
Our ICC formulation examines this distinction for conditional population responses: background variation can remain while matched treatment--control contrasts disappear.

\paragraph{Learning change through difference targets.}
Concurrent work MotionJEPA~\cite{karmann2026motionjepa} learns temporal visual dynamics from consecutive images and evaluates downstream planning.
Its DISReg objective predicts temporal-difference image embeddings from paired state embeddings, regularizing both to penalize constant targets.
PerturbJEPA instead models cellular perturbation responses under destructive single-cell measurement, with unpaired control and treated populations, and retrieves the source drug--dose intervention from an observed response.
Our difference target uses mean-expression contrasts without cell correspondence, and PEA aligns it with a fixed scaled centroid displacement rather than a learned difference predictor.
Thus, whereas MotionJEPA makes temporal change recoverable from paired visual states, PEA specifies where population response must appear in the geometry used for retrieval.
Our finite-batch analysis relates retained effect-target variation and alignment error to measured centroid displacements.

\paragraph{Population response prediction.}
scGen represents perturbation effects through differences between latent population means~\cite{lotfollahi2019scgen}; CPA composes basal states, perturbations, and covariates~\cite{lotfollahi2023cpa}; and CellOT learns mappings between unpaired populations using optimal transport~\cite{bunne2023cellot}.
STATE combines cell-state representations with conditional transitions and decodes predicted states into expression~\cite{adduri2026state}.
We shape the predicted embedding space for direct response matching, comparing candidate predictions with the observed target in that space.

\paragraph{Cellular JEPA models.}
Cell-JEPA combines masked-view latent prediction with expression reconstruction, then uses perturbation-label embeddings and a perturbed-state JEPA objective for response prediction~\cite{elsheikh2026celljepa}.
Its distinction between absolute-state prediction and effect-size estimation motivates assessing response information separately from state quality.
BioJEPA-AC also conditions latent prediction on genetic and chemical interventions~\cite{jemec2026biojepa}.
JEPA-Anything predicts latent treatment--control residuals through factorized pathways and evaluates decoded expression~\cite{cui2026jepaanything}.
Building on intervention-conditioned latent prediction, we focus on explicit population-displacement constraints and response-based candidate ranking.

\paragraph{Response-based intervention retrieval.}
The Connectivity Map and L1000 connect interventions through observed expression signatures~\cite{lamb2006connectivity,subramanian2017l1000}.
GeneSpeak-FP learns compound and target retrieval from observed cell-level responses relative to control, using a closed candidate library and held-out drug--cell-line groups~\cite{vaniushkina2026genespeak}.
PerturbJEPA instead predicts each candidate response from a common control population and compares these predictions with an observed target.
This forward-model interface supports a shared retrieval task across systems with different output spaces.
Complementing Systema's emphasis on perturbation-specific effects beyond systematic variation~\cite{systema2025}, our evaluation tests whether predictions distinguish competing interventions within a common context.

\section{Method}
\label{sec:method}

PerturbJEPA learns cell population responses for \emph{source-intervention retrieval}: given a control population and an observed target response, it ranks a catalog of drug--dose actions to recover the recorded source. Figure~\ref{fig:architecture} summarizes the model. Population prediction, effect preservation, and action discrimination impose complementary requirements on its learned representation.

\subsection{Population prediction and intervention retrieval}
\label{sec:formulation}

A training example contains unpaired control and treated cell populations $X_b^0,X_b^1\in\mathbb R^{n\times G}$ and action $a_b=(d_b,s_b)$. Controls match the treated population's plate and cellular background. As shown in Fig.~\ref{fig:architecture}(a), a shared encoder $E_\theta$, intervention encoder $e_\eta$, and predictor $P_\varphi$ produce
\begin{equation}
 Z_b^r=E_\theta(X_b^r),\qquad
 \widehat Z_b(a)=P_\varphi(Z_b^0,e_\eta(a)),
 \label{eq:forward}
\end{equation}
where $r\in\{0,1\}$ and $Z_b^r\in\mathbb R^{n\times D}$. Expression counts are library-normalized to $10^4$, log-transformed, and restricted to $G=2{,}000$ highly variable genes. Each cell starts from the token $t(x)=\sum_gx_gE_g/\|x\|_2$, with a learned token for zero expression. The denominator normalizes expression weights, not the resulting embedding. An eight-layer set transformer ($D=512$, eight heads, feed-forward width $2{,}048$) uses RMSNorm and SwiGLU without positional encoding; attention operates only within each population.

The action encoder constructs a learned drug token and a dose token from a $1\!\to\!512\!\to\!512$ SiLU MLP applied to $\log(1+s/s_{\min}^{+})$, where $s_{\min}^{+}$ is the smallest positive vocabulary dose. Both receive type embeddings and RMSNorm. Their mean conditions four predictor blocks through adaptive RMSNorm shifts, scales, and residual gates, initialized to zero; a final affine RMSNorm produces the output. Drug identities use a lookup table, restricting prediction to known compounds.

\paragraph{Population Energy Discrepancy (PED).}
Unpaired cells require distributional prediction. For populations $U=\{u_i\}_{i=1}^m$, $V=\{v_j\}_{j=1}^n$ and $\rho(u,v)=\|u-v\|_2/\sqrt D$, define
\begin{equation}
\begin{split}
 \mathcal D_E(U,V)={}&\frac1{mn}\sum_{i,j}\rho(u_i,v_j)
 -\frac1{2m^2}\sum_{i,i'}\rho(u_i,u_{i'})\\
 &-\frac1{2n^2}\sum_{j,j'}\rho(v_j,v_{j'}).
\end{split}
\label{eq:energy}
\end{equation}
This empirical V-statistic includes diagonal terms and equals half the conventional energy distance~\cite{szekely2013energy}. Its within-population terms retain sensitivity beyond mean matching. We minimize $\mathcal L_{\mathrm{PED}}=B^{-1}\sum_b\mathcal D_E(\widehat Z_b(a_b),Z_b^1)$ over a synchronized batch of $B$ pairs. Both state branches receive gradients; there is no momentum teacher or stopped target gradient. The discrepancy compares realized empirical measures and requires no independence among encoded slots.

At inference, encode the query once as $Z^0=E_\theta(X^0)$ and $Z^\star=E_\theta(X^\star)$. Each candidate predicts from the same control:
\begin{equation}
 \widehat a=\operatorname*{arg\,min}_{a\in\mathcal A}
 \mathcal D_E\!\left(P_\varphi(Z^0,e_\eta(a)),Z^\star\right).
 \label{eq:inverse}
\end{equation}
The catalog contains the source action. Full populations are scored in raw latent coordinates; drug-level retrieval selects each drug's best-scoring dose. The effect branch described below is used only during training.

\subsection{Interventional contrast collapse}
\label{sec:icc}

PED admits an all-constant solution. \textbf{State-Marginal Isotropy (SMI)} discourages this global collapse using sample-count-normalized SIGReg~\cite{balestriero2025lejepa}. Write $\mathcal I(U)$ for its discrepancy between projected empirical characteristic functions and a standard Gaussian. We use
\begin{equation}
 \mathcal R_{\mathrm{SMI}}=\frac1{2n}\sum_{r\in\{0,1\}}\sum_{i=1}^n
 \mathcal I\!\left(\{z_{bi}^r\}_{b=1}^B\right).
 \label{eq:smi}
\end{equation}
Each slot is a sampled position, not a correspondence between cells. Statistics are aggregated across workers, separately for control and treated states. We use $256$ random unit projections and $17$ frequencies in $[0,3]$, without centering or standardizing inputs.

Global dispersion does not ensure preservation of intervention responses. Fix a background, an intervention, and a population-sampling protocol. \textbf{Interventional Contrast Collapse (ICC)} occurs when control and treated expression populations have distinct laws but their \emph{encoded empirical measures} have the same law, while representations remain globally nondegenerate. The definition concerns distributions of sampled populations, including dependencies induced by set encoding.

An illustrative construction exposes the gap. For input populations containing $n$ repeated observations $(H,R^r)$, let the encoder map every slot to
\begin{equation}
 z_s^r=\sqrt{1-s^2}H+sR^r,\qquad 0\leq s\leq1,
 \label{eq:contraction}
\end{equation}
where $H\sim\mathcal N(0,I_D)$ is shared by the pair, each $R^r$ is marginally standard Gaussian, and $H$ is independent of $(R^0,R^1)$. Every state marginal remains standard Gaussian. Yet a predictor copying the control incurs expected PED $s\,\mathbb E\|R^1-R^0\|_2/\sqrt D$: within-population distances vanish and all cross-distances scale with $s$. Shrinking $s$ reduces prediction error while preserving state marginals. For a genuine conditional response, take $R^0=U,R^1=A$ with independent Gaussian $H,U,A$, and condition on background $H$ and action $A$. The input laws differ, but at $s=0$ both encoded empirical measures equal $\delta_H$. This construction demonstrates that ICC can coexist with nondegenerate state marginals; Figure~\ref{fig:mechanism} examines contrast contraction during training.

\subsection{Preserving effects in population geometry}
\label{sec:pea}

\textbf{Pseudobulk-to-latent Effect Alignment (PEA)} constrains the measured population change (Fig.~\ref{fig:architecture}(b)). Let $\bar x_b^r$ and $\bar z_b^r$ be population means, $\Delta x_b=\bar x_b^1-\bar x_b^0$, and $\Delta z_b=\bar z_b^1-\bar z_b^0$. An independent effect encoder $g$ yields
\begin{equation}
 v_b=g(\Delta x_b),\qquad
 \mathcal L_{\mathrm{PEA}}=\frac1{BD}\sum_b
 \left\|\frac{\Delta z_b}{\alpha}-v_b\right\|_2^2.
 \label{eq:pea}
\end{equation}
We set $\alpha=0.2$ and use the bias-free network $g(x)=W_2\tanh(W_1(16x))$ with widths $2{,}000\!\to\!128\!\to\!512$, without output normalization. It satisfies $g(0)=0$ and $g(-x)=-g(x)$. Gradients update both state branches and the effect encoder. The fixed displacement operator directly places the learned effect in the centroid geometry of unpaired populations.

Alignment alone permits $\Delta z_b=v_b=0$. We therefore add \textbf{Effect-Space Isotropy (ESI)}, $\mathcal R_{\mathrm{ESI}}=\mathcal I(\{v_b\}_{b=1}^B)$, across unscaled effect targets. A constant batch incurs at least SIGReg's positive all-zero penalty: for each projected frequency, $|e^{iu}-\phi|^2\geq(1-\phi)^2$ with Gaussian characteristic value $\phi>0$. The bottleneck limits effects to at most $128$ dimensions, so ESI regularizes projected variation rather than attaining full $512$-dimensional isotropy.

\Needspace{8\baselineskip}
\begin{proposition}[Finite-batch contrast transfer]
\label{prop:contrast-transfer}
Let $V,\Delta Z\in\mathbb R^{B\times D}$ stack the effect targets and latent displacements, and let $H_B=I_B-\mathbf1\mathbf1^\top/B$. Define $S_v=\|H_BV\|_F/\sqrt{BD}$, $S_\Delta=\|H_B\Delta Z\|_F/\sqrt{BD}$, and $\epsilon=\mathcal L_{\mathrm{PEA}}$. Then
\begin{equation}
 S_\Delta\geq\alpha\bigl(S_v-\sqrt\epsilon\bigr)_+.
 \label{eq:pea-main-bound}
\end{equation}
\end{proposition}
\begin{proof}
Equation~\eqref{eq:pea} gives $\|\Delta Z/\alpha-V\|_F=\sqrt{BD\epsilon}$. Since $H_B$ is an orthogonal projection, centering cannot increase this norm. Applying the reverse triangle inequality to the centered matrices and dividing by $\sqrt{BD}$ proves the bound.
\end{proof}

Proposition~\ref{prop:contrast-transfer} shows that when variation in the effect targets exceeds the root-mean-square alignment error, the measured centroid displacements cannot all be equal within a batch. These displacements could still vary across cellular backgrounds while remaining identical for every intervention within each background. To distinguish candidates within the same background, we use IDR to train the predictor to rank the source drug above other drugs at the same dose and the source dose above other doses of the same drug.

\subsection{Intervention-discriminative ranking}
\label{sec:idr}

\textbf{Intervention-Discriminative Ranking (IDR)} compares all candidate predictions for a query against the same encoded treated population (Fig.~\ref{fig:architecture}(c)). After encoding each full population, take its first $R=32$ slots, denoted $Z_b^{r,[R]}$, and rerun the predictor on the reduced control for every candidate, including the positive:
\begin{equation}
\begin{aligned}
 q_b(a)&=-\mathcal D_E\!\left(P_\varphi(Z_b^{0,[R]},e_\eta(a)),Z_b^{1,[R]}\right)/\tau,\\
 \ell_b(\mathcal C)&=-\log\frac{\exp q_b(a_b)}{\sum_{a\in\mathcal C}\exp q_b(a)},
\end{aligned}
\label{eq:idr}
\end{equation}
with $\tau=0.01$. Subsets are compared as distributions, without cell matching. The drug candidate set contains the source and eight different drugs sharing its dose identifier, selected by highest cosine similarity between normalized mean action tokens. Mining is detached, then selected actions are re-encoded with gradients. The dose candidate set contains the source and the other two observed dose identifiers for that drug. Numerical dose encoding is continuous, but candidate eligibility uses vocabulary identifiers.

Averaging $\ell_b$ over these sets gives $\mathcal L_{\mathrm{IDR}}^{\mathrm{drug}}$ and $\mathcal L_{\mathrm{IDR}}^{\mathrm{dose}}$. They share the positive score but use separate nine-way and three-way softmaxes; no monotonic dose response is imposed. Gradients traverse the action encoder, predictor, and both state branches. The full objective is
\begin{equation}
\begin{split}
 \mathcal L={}&\mathcal L_{\mathrm{PED}}+\mathcal R_{\mathrm{SMI}}
 +0.25\mathcal L_{\mathrm{PEA}}+\mathcal R_{\mathrm{ESI}}\\
 &+10^{-3}\mathcal L_{\mathrm{IDR}}^{\mathrm{drug}}
 +10^{-3}\mathcal L_{\mathrm{IDR}}^{\mathrm{dose}}.
\end{split}
\label{eq:total}
\end{equation}

\section{Intervention retrieval experiments}
\label{sec:experiments}

\paragraph{Evaluation protocol.}
We evaluate source-intervention recovery on Tahoe-100M~\citep{zhang2026tahoe} using $3{,}592$ held-out drug--dose--background queries spanning five cell lines.
Each query contains $256$ distinct treated cells and $256$ matched control cells with the same plate composition.
We exclude $81$ test groups with fewer than $256$ treated cells; their actions can remain candidates.
Each query ranks its complete background-specific catalog of $734$ or $735$ drug--dose actions from $268$ drugs.
The split holds out combinations of known interventions and cellular backgrounds; it does not test unseen drugs or unseen cell lines.
All methods receive the same frozen cells and candidate catalogs, with no test-time parameter updates.

Joint R@1 measures recovery of the exact source drug--dose action.
Drug R@1 ranks each drug by its best-scoring available dose.
Conditional dose accuracy selects among the true drug's doses for every query, regardless of drug-ranking accuracy.
We use a hierarchical macro average: doses within drugs, drugs within backgrounds, and then the five backgrounds with equal weight.
Recall and accuracy are percentages; MRR is multiplied by $100$.
This weighting differs from pooling all requests: the full model's joint R@1 is $91.46\%$ macro versus $91.31\%$ request-weighted.

\paragraph{Systems, selection, and uncertainty.}
We compare Mean, additive ridge Linear, CPA~\citep{lotfollahi2023cpa}, the official STATE checkpoint~\citep{adduri2026state}, both official PerturbDiff Tahoe releases~\citep{yuan2026perturbdiff}, and a CellFlow adaptation~\citep{klein2025cellflow}.
Mean and Linear predict training-derived expression means; CPA is adapted from the official implementation.
CellFlow retains the official flow-matching model and solver, with a training-only 100-dimensional PCA projection and a fixed training budget.
Energy discrepancy scores every system in its own prediction space: expression for the first six rows of Table~\ref{tab:main-retrieval}, PCA for CellFlow, and learned 512-dimensional populations for PerturbJEPA.
For each PerturbJEPA configuration, validation joint R@1 selects a checkpoint on a common $263$-query panel, with joint MRR breaking ties, before test evaluation.
Checkpoint selection uses a common catalog of $379$ drugs at three positive doses ($1{,}137$ actions per background).
We evaluate saved checkpoints every $10$k updates.
Each configuration uses one training seed ($42$).
Paired $95\%$ confidence intervals (CIs) use $4{,}000$ target-drug-cluster bootstrap replicates, retaining linked doses and backgrounds.
They condition on the fitted checkpoints and frozen cells, and do not estimate variation across training seeds or cell resampling.

\paragraph{Implementation.}
The shared-encoder recipe uses $240$ population pairs per update, $256$ cells per population, AdamW with peak learning rate $5\times10^{-5}$ and weight decay $10^{-3}$, gradient clipping at $1$, and BF16.
The training budget is $200{,}000$ updates, with $10{,}000$ warmup updates followed by cosine decay.

\begin{table}[t]
\centering
\small
\setlength{\tabcolsep}{2pt}
\caption{\textbf{Overall intervention retrieval.} All eight systems use the same 3,592 queries and hierarchical macro aggregation. R@1 and dose accuracy are percentages; MRR is $\times100$. Dose accuracy supplies the true drug.}
\label{tab:main-retrieval}
\begin{tabular*}{\columnwidth}{@{\extracolsep{\fill}}lrrrr@{}}
\toprule
& \multicolumn{2}{c}{Joint} & Drug & Dose\\
\cmidrule(lr){2-3}
Model & R@1 & MRR & R@1 & Acc.\\
\midrule
Mean & 4.47 & 9.24 & 5.90 & 61.52\\
Linear & 75.57 & 82.94 & 78.16 & 95.99\\
CPA & 0.10 & 0.88 & 0.30 & 36.58\\
PerturbDiff (scratch) & 1.92 & 4.17 & 2.90 & 49.38\\
PerturbDiff (fine-tuned) & 2.93 & 5.96 & 3.42 & 48.70\\
STATE & 88.06 & 91.39 & 89.44 & 97.36\\
CellFlow & 4.63 & 11.78 & 5.86 & 76.94\\
\midrule
\textbf{PerturbJEPA} & \textbf{91.46} & \textbf{94.43} & \textbf{91.74} & \textbf{99.31}\\
\bottomrule
\end{tabular*}
\end{table}

\begin{table*}[t]
\centering
\small
\setlength{\tabcolsep}{3pt}
\renewcommand{\arraystretch}{1.08}
\caption{\textbf{Effect preservation and intervention discrimination.} Panels A and B share reference (d). Panel C tests PEA with momentum targets. Panel D replaces only the PEA alignment in the full model (e). All rows report test results at validation-selected checkpoints.}
\label{tab:ablation-ranking}
\begin{tabular*}{\textwidth}{@{\extracolsep{\fill}}clccccccrrrr@{}}
\toprule
\multirow{2}{*}{Row} & \multirow{2}{*}{Configuration} & \multicolumn{6}{c}{Enabled training objectives} & \multicolumn{2}{c}{Joint} & Drug & Dose\\
\cmidrule(lr){3-8}\cmidrule(lr){9-10}\cmidrule(lr){11-11}\cmidrule(l){12-12}
& & PED & SMI & PEA & ESI & \shortstack{Drug\\IDR} & \shortstack{Dose\\IDR} & R@1 & MRR & R@1 & Acc.\\
\midrule
\multicolumn{12}{@{}l}{\textbf{A. Preserve intervention effects}\quad\textit{Shared encoder; drug IDR fixed}}\\
(a) & No effect branch & $\checkmark$ & $\checkmark$ & $\times$ & $\times$ & $\checkmark$ & $\times$ & 1.02 & 2.36 & 1.92 & 37.89\\
(b) & + PEA alignment & $\checkmark$ & $\checkmark$ & $\delta$ & $\times$ & $\checkmark$ & $\times$ & 18.57 & 29.79 & 20.72 & 84.18\\
(d) & + ESI (reference) & $\checkmark$ & $\checkmark$ & $\delta$ & $\checkmark$ & $\checkmark$ & $\times$ & 87.47 & 92.07 & 89.83 & 96.47\\
\addlinespace[3pt]
\multicolumn{12}{@{}l}{\textbf{B. Discriminate drugs and doses}\quad\textit{Shared encoder; complete PEA fixed}}\\
(c) & No ranking losses & $\checkmark$ & $\checkmark$ & $\delta$ & $\checkmark$ & $\times$ & $\times$ & 82.45 & 88.49 & 85.12 & 95.97\\
(d) & + Drug IDR (reference) & $\checkmark$ & $\checkmark$ & $\delta$ & $\checkmark$ & $\checkmark$ & $\times$ & 87.47 & 92.07 & 89.83 & 96.47\\
(e) & \textbf{+ Dose IDR (full)} & $\checkmark$ & $\checkmark$ & $\delta$ & $\checkmark$ & $\checkmark$ & $\checkmark$ & \textbf{91.46} & \textbf{94.43} & \textbf{91.74} & \textbf{99.31}\\
\addlinespace[3pt]
\multicolumn{12}{@{}l}{\textbf{C. Effect preservation with momentum targets}\quad\textit{Both IDR terms fixed}}\\
(f) & EMA, no effect branch & $\checkmark$ & $\times$ & $\times$ & $\times$ & $\checkmark$ & $\checkmark$ & 4.53 & 9.90 & 7.33 & 55.00\\
(g) & EMA + complete PEA & $\checkmark$ & $\times$ & $\delta$ & $\checkmark$ & $\checkmark$ & $\checkmark$ & 77.00 & 84.26 & 79.55 & 95.55\\
\addlinespace[3pt]
\multicolumn{12}{@{}l}{\textbf{D. Direct versus learned effect alignment}\quad\textit{Full model (e) as reference}}\\
(h) & Learned endpoints & $\checkmark$ & $\checkmark$ & $h_E$ & $\checkmark$ & $\checkmark$ & $\checkmark$ & 81.73 & 87.84 & 82.23 & 98.24\\
(i) & Learned displacement & $\checkmark$ & $\checkmark$ & $h_\Delta$ & $\checkmark$ & $\checkmark$ & $\checkmark$ & 58.18 & 70.76 & 60.18 & 95.05\\
\bottomrule
\end{tabular*}
\par\smallskip
\parbox{\textwidth}{\small
$\checkmark$ and $\times$ denote active and inactive losses. PEA uses direct displacement ($\delta$), an endpoint head ($h_E$), or a displacement head ($h_\Delta$). Weights are $1$ for PED, SMI, and ESI, $0.25$ for PEA, and $10^{-3}$ for each IDR term. Metric units follow Table~\ref{tab:main-retrieval}.}
\end{table*}

\paragraph{Overall retrieval.}
PerturbJEPA achieves $91.46\%$ joint R@1 and $94.43$ MRR, compared with STATE's $88.06\%$ and $91.39$.
The paired gains are $3.40$ percentage points in joint R@1 (CI $[1.72,5.07]$) and $3.04$ MRR units ($[1.80,4.29]$).
Drug R@1 improves by $2.29$ points ($[0.62,3.90]$), and conditional dose accuracy reaches $99.31\%$.
Linear attains $75.57\%$ joint R@1, showing that additive training-set responses already contain substantial retrieval information.

\subsection{Controlled contrast diagnostics}
\label{sec:mechanism}

\begin{figure*}[t]
\centering
\includegraphics[width=\textwidth]{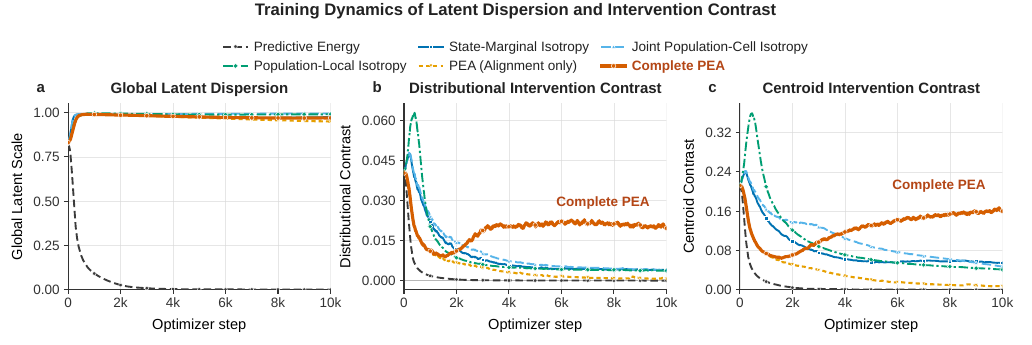}
\caption{\textbf{Global diversity and intervention contrast require distinct constraints.}
Six controlled runs share common initial weights and actual sampled training batches.
(a) Global centered latent scale; (b) energy discrepancy between encoded treated and control populations; (c) their dimension-normalized centroid distance.
State regularization sustains global scale but permits contrast contraction; complete PEA sustains larger contrasts than alignment alone.
Both IDR terms are disabled for all 10,000 warmup updates.
Curves are centered 101-step moving means from one training seed, with no uncertainty bands.
Diagnostics use observed training populations, include sampling variation, and are not prediction errors.}
\label{fig:mechanism}
\end{figure*}

Figure~\ref{fig:mechanism} isolates representation behavior before ranking supervision.
The state-only controls apply SIGReg across population pairs at each cell slot, across pooled population--cell slots, or separately within each population.
With $60$ population pairs and $256$ cells, their respective sample counts are $60$, $15{,}360$, and $256$; all use the same sample-count-normalized penalty.
This comparison tests whether changing the axis of state regularization preserves intervention contrasts while PED remains fixed.
The alignment-only and complete-PEA configurations instead retain population-axis SMI, isolating the added effect constraints.
PED alone collapses globally, whereas all three state-regularization axes retain scale near one while treatment--control contrasts contract.
Alignment alone also leaves much smaller contrasts than complete PEA.
Over updates 9,001--10,000, complete PEA gives mean energy and centroid contrasts of $0.0202$ and $0.1599$, compared with $0.00402$ and $0.0562$ for SMI alone, at comparable global scales ($0.9697$ and $0.9737$).
These trajectories measure treatment--control separation. Two interventions can induce equally large shifts, so held-out retrieval tests whether the representation also distinguishes their responses.

\subsection{Effect preservation and candidate discrimination}
\label{sec:ablations}
\label{sec:ablation-ranking}

\paragraph{Preserving intervention effects.}
With drug IDR fixed (Table~\ref{tab:ablation-ranking}A), joint R@1 is $1.02\%$ without the effect branch. PEA alignment raises it to $18.57\%$, a gain of $17.55$ points ($[15.83,19.28]$).
Adding ESI completes PEA and raises retrieval to $87.47\%$, a further $68.90$ points ($[66.75,71.02]$).
Joint MRR follows the same progression, from $2.36$ to $29.79$ to $92.07$.
The larger gain from ESI shows the value of maintaining variation in the effect targets.

\paragraph{Discriminating drugs and doses.}
With complete PEA fixed, the configuration without either IDR term reaches $82.45\%$ joint R@1.
Adding drug IDR improves this by $5.01$ points ($[3.84,6.19]$), and drug R@1 rises from $85.12\%$ to $89.83\%$.
Adding dose IDR then improves joint R@1 by $4.00$ points ($[3.01,5.02]$) to $91.46\%$.
Conditional dose accuracy rises from $96.47\%$ to $99.31\%$, while mean absolute $\log_{10}$ dose error falls from $0.0495$ to $0.0093$.
Both configurations receive numerical doses, isolating the gain from within-drug ranking supervision.
Joint R@50 changes only from $99.15\%$ without IDR to $99.71\%$ with both terms.

\paragraph{Momentum targets.}
We also test effect preservation within an online/EMA-target JEPA recipe that omits SMI and retains both IDR terms.
Adding complete PEA increases joint R@1 from $4.53\%$ to $77.00\%$, a paired gain of $72.46$ points ($[70.28,74.54]$).
PEA computes both centroids with the online encoder, while prediction and ranking use stop-gradient EMA targets.

\paragraph{Direct versus learned effect alignment.}
In concurrent work, MotionJEPA~\cite{karmann2026motionjepa} predicts an encoded frame difference from two consecutive frame embeddings.
To test this design for intervention retrieval, row (h) predicts the pseudobulk effect from the control and treated centroids of unpaired populations.
Row (i) gives the head their scaled difference, isolating the effect of adding a learned map to direct PEA (e).
With all six losses active, joint R@1 is $91.46\%$ for direct PEA, $81.73\%$ for the endpoint head, and $58.18\%$ for the displacement head.
The gap for the endpoint head is already evident in drug R@1 ($91.74\%$ to $82.23\%$), while conditional dose accuracy changes by only $1.07$ points.

\subsection{Design choices and input coverage}
\label{sec:ablation-design}

With drug IDR as the sole ranking objective, we vary how expression weights are normalized, how the intervention conditions the predictor, and how population discrepancy is measured during training and retrieval.
We then compare the full model trained on $2{,}000$ highly variable genes with its all-gene counterpart.

\begin{table}[t]
\centering
\small
\setlength{\tabcolsep}{3pt}
\renewcommand{\arraystretch}{1.15}
\caption{\textbf{Expression-weight normalization.} For $x\ne0$, cell tokens are $t(x)=\sum_g x_g E_g/d(x)$. Other settings follow the drug-IDR reference in Table~\ref{tab:ablation-ranking}(d).}
\label{tab:ablation-normalization}
\begin{tabular*}{\columnwidth}{@{\extracolsep{\fill}}lrrrr@{}}
\toprule
\multirow{2}{*}{Denominator $d(x)$} & \multicolumn{2}{c}{Joint} & Drug & Dose\\
\cmidrule(lr){2-3}\cmidrule(lr){4-4}\cmidrule(l){5-5}
& R@1 & MRR & R@1 & Acc.\\
\midrule
$\|x\|_2$ & 87.47 & 92.07 & 89.83 & 96.47\\
$1$ & 87.55 & 91.99 & 89.99 & 96.43\\
$\|x\|_1$ & 38.72 & 52.29 & 42.47 & 86.96\\
\bottomrule
\end{tabular*}
\end{table}

\paragraph{Expression-weight normalization.}
In Table~\ref{tab:ablation-normalization}, removing $\ell_2$ normalization changes joint R@1 by only $+0.09$ points ($[-1.18,1.38]$), whereas $\ell_1$ normalization reaches $38.72\%$.
Each denominator rescales the same weighted gene sum, preserving relative gene weights within a cell.
The drop under $\ell_1$ points to sensitivity to input scale, while the unnormalized sum performs comparably to the $\ell_2$ reference.

\begin{table}[t]
\centering
\small
\setlength{\tabcolsep}{3pt}
\renewcommand{\arraystretch}{1.15}
\caption{\textbf{Intervention conditioning.} Depth, width, and intervention encodings are matched. Both variants retain the reference input normalization and training objective from Table~\ref{tab:ablation-ranking}(d).}
\label{tab:ablation-conditioning}
\begin{tabular*}{\columnwidth}{@{\extracolsep{\fill}}lrrrr@{}}
\toprule
\multirow{2}{*}{Conditioning} & \multicolumn{2}{c}{Joint} & Drug & Dose\\
\cmidrule(lr){2-3}\cmidrule(lr){4-4}\cmidrule(l){5-5}
& R@1 & MRR & R@1 & Acc.\\
\midrule
AdaLN & 87.47 & 92.07 & 89.83 & 96.47\\
Prefix tokens & 73.64 & 81.80 & 78.17 & 92.50\\
\bottomrule
\end{tabular*}
\end{table}

\begin{table}[t]
\centering
\small
\setlength{\tabcolsep}{2pt}
\caption{\textbf{Retrieval-distance sensitivity.} The two independently validation-selected checkpoints are fixed while changing the test scorer. Both use complete PEA and drug IDR, with dose IDR disabled, and retain their original energy-based validation selection.}
\label{tab:ablation-distance-sensitivity}
\begin{tabular*}{\columnwidth}{@{\extracolsep{\fill}}lrrrr@{}}
\toprule
& \multicolumn{2}{c}{Energy scoring} & \multicolumn{2}{c}{Sinkhorn scoring}\\
\cmidrule(lr){2-3}\cmidrule(l){4-5}
Training & R@1 & MRR & R@1 & MRR\\
\midrule
Energy & 87.47 & 92.07 & 86.74 & 91.51\\
Sinkhorn & 62.89 & 73.39 & 63.07 & 73.50\\
\bottomrule
\end{tabular*}
\par\smallskip
{\small Joint R@1 is a percentage; joint MRR is $\times100$.}
\end{table}

\paragraph{Intervention conditioning.}
Replacing AdaLN with prefix tokens lowers joint R@1 to $73.64\%$ (Table~\ref{tab:ablation-conditioning}).
Both predictors receive the same intervention encodings: AdaLN modulates every block, while the prefix model incorporates them through full self-attention over intervention and cell tokens.
The stronger AdaLN result supports adaptive modulation for converting intervention information into predicted responses in PerturbJEPA.

\paragraph{Training and retrieval discrepancy.}
Replacing energy discrepancy with Sinkhorn divergence in both population prediction and drug ranking yields $62.89\%$ joint R@1 under energy scoring, with auxiliary weights and ranking temperature retained.
Rescoring the fixed checkpoints with Sinkhorn gives $86.74\%$ for energy training and $63.07\%$ for Sinkhorn training (Table~\ref{tab:ablation-distance-sensitivity}).
Energy training retains its advantage under both scorers.
The much smaller changes from rescoring show that the training recipe has a larger effect than the test discrepancy in this comparison.

\paragraph{Input coverage.}
Table~\ref{tab:ablation-gene-coverage} expands the complete model from $2{,}000$ HVGs to all $62{,}710$ genes and raises joint R@1 to $94.50\%$, a paired gain of $3.04$ points ($[1.57,4.56]$), and joint MRR to $96.81$.
Drug R@1 reaches $96.30\%$, whereas conditional dose accuracy decreases to $97.72\%$.
The expanded gene-embedding and effect-input matrices add both expression coverage and input-layer capacity, with hidden width, depth, and optimization settings held fixed.
The gain favors drug identification, while the decline in conditional dose accuracy shows that greater input coverage does not automatically sharpen within-drug dose resolution.

The supplementary material provides complete theoretical analysis and proofs, detailed ablation settings, baseline configurations, checkpoint selection, and additional aggregate retrieval metrics.

\begin{table}[t]
\centering
\small
\setlength{\tabcolsep}{2pt}
\caption{\textbf{Input-gene coverage.} Both models use all six objectives and their own validation-selected checkpoints. Hidden width, depth, and optimization settings are matched; expanded input matrices increase parameter count. Units follow Table~\ref{tab:main-retrieval}.}
\label{tab:ablation-gene-coverage}
\begin{tabular*}{\columnwidth}{@{\extracolsep{\fill}}lrrrrr@{}}
\toprule
& & \multicolumn{2}{c}{Joint} & Drug & Dose\\
\cmidrule(lr){3-4}\cmidrule(lr){5-5}\cmidrule(l){6-6}
Input & Genes & R@1 & MRR & R@1 & Acc.\\
\midrule
HVG2000 & 2,000 & 91.46 & 94.43 & 91.74 & 99.31\\
All genes & 62,710 & 94.50 & 96.81 & 96.30 & 97.72\\
\bottomrule
\end{tabular*}
\end{table}

\section{Conclusion and limitations}
\label{sec:conclusion}
The controlled trajectories show that an encoder can retain global diversity as the distance between treated and control populations shrinks.
PerturbJEPA uses the change in mean expression to supervise the displacement between population centroids. The finite-batch bound relates variation in the effect targets to variation in these displacements when alignment error is small.
Both learned heads predict the same type of effect target, yet joint recall is highest when the loss acts directly on the centroid displacement in the space used for retrieval.
Drug and dose ranking train the predictor to distinguish candidate responses from a common control. Together these objectives yield $91.46\%$ macro joint Recall@1 on Tahoe-100M, compared with $88.06\%$ for STATE.

Two treatments can produce the same mean shift with different mixtures of responding cells. They then supply the same target to PEA. Distinguishing them depends on whether the full population embeddings retain the distributional difference scored by PED.

Evaluation holds out combinations of known drugs and cellular backgrounds. Extending retrieval to unseen compounds would require replacing the drug lookup, for example with a chemical-structure encoder. Transfer to unseen cell lines remains open.

\onecolumn
\begin{multicols}{2}
\phantomsection
\addcontentsline{toc}{section}{References}
{\small
  \setlength{\bibsep}{4pt plus 1pt minus 1pt}
  \bibliographystyle{ieeenat_fullname}
  \bibliography{references}
}
\end{multicols}

\appendix
\twocolumn[{\centering\Large\bfseries Supplementary Material\par\vspace{12pt}}]
\section{Implementation and controlled diagnostics}
\label{app:implementation}

\paragraph{Input and architecture.}
\label{app:input-architecture}
For raw counts $c_g$, we compute $x_g=\log(1+10^4c_g/\sum_{g'}c_{g'})$ over the full library before retaining the fixed 2,000 highly variable genes (HVGs).
The all-gene comparison instead retains 62,710 genes.
The expression-weighted token $\sum_gx_gE_g/\|x\|_2$ uses a learned empty-cell token when the denominator is zero.
This normalizes expression weights, not the resulting embedding vector.
The population encoder has eight layers, eight heads, width 512, feed-forward width 2,048, RMSNorm, and SwiGLU.
There is no positional encoding, causal mask, or attention between observed control and treated populations.

Drug identity uses a learned lookup embedding and a learned type embedding followed by RMSNorm.
Dose uses a $1\to512\to512$ SiLU MLP on $\log(1+s/s_{\min}^{+})$, followed by a dose-type embedding and RMSNorm; $s_{\min}^{+}$ is the smallest positive vocabulary dose.
The mean of these two tokens controls shifts, scales, and residual gates in a four-layer predictor.
Adaptive blocks use non-affine RMSNorm and a final affine RMSNorm; modulation and gates are zero-initialized.
The effect encoder is $g(\Delta x)=W_2\tanh(W_1(16\Delta x))$, without biases or output normalization, with widths $2000\to128\to512$ in the primary model; the all-gene model changes the first width to 62,710.
Thus $g(0)=0$ and $g(-\Delta x)=-g(\Delta x)$.
Its output lies in a subspace of dimension at most 128, and ESI regularizes its projected variation.

\paragraph{Exact regularizer.}
\label{app:regularizer}
For vectors $U=\{u_m\}_{m=1}^M$, draw $K=256$ unit Gaussian directions $a_k$ and use $L=17$ frequencies $t_\ell=\ell h$, $h=3/16$.
Let $\phi_\ell=\exp(-t_\ell^2/2)$, with weights $q_0=q_{16}=h$ and $q_\ell=2h$ otherwise.
The implemented sample-count-normalized SIGReg~\citep{balestriero2025lejepa} is
\begin{equation}
\mathcal I(U)=\frac1K\sum_{k,\ell}q_\ell\phi_\ell
\left|\frac1M\sum_m e^{it_\ell a_k^\top u_m}-\phi_\ell\right|^2.
\end{equation}
Inputs are not centered or standardized.
The quadrature weights include negative frequencies, and there is no additional sample-count factor.
Directions are synchronized across workers and refreshed across calls; characteristic functions are aggregated over all workers.
SMI applies this penalty separately to each state and cell slot across population pairs.
ESI applies it across population-level effect embeddings.
Section~\ref{supp:theory:theory-esi} gives the corresponding finite-batch analysis.

\paragraph{Training and ranking.}
\label{app:training-ranking}
The main-paper PerturbJEPA recipes use training seed 42, 256 cells per population, 15 population pairs per worker, AdamW with peak learning rate $5\times10^{-5}$, weight decay $10^{-3}$, gradient clipping at 1, and BF16 without gradient accumulation.
The PerturbJEPA retrieval runs use a 200,000-update schedule with 10,000 warmup updates followed by cosine decay.
The shared-encoder retrieval suite uses 16 workers (240 pairs per update); the EMA suite and controlled diagnostics use four workers (60 pairs).
After encoding full populations, ranking selects the first 32 slots and reruns the predictor on that reduced control set for every candidate, including the positive.
With both IDR terms enabled, one positive, eight same-dose alternative drugs, and two same-drug alternative doses are scored together, then partitioned into separate nine-way and three-way cross-entropies at temperature 0.01.
Hard drug neighbors are chosen without gradients by cosine similarity of normalized mean intervention tokens, then re-encoded with gradients.
The two positive-dose alternatives are required by the Tahoe training vocabulary; vehicle controls are not negative candidates.

For energy-based training, PED and ranking use float32 tensorized GeomLoss~\citep{feydy2019sinkhorn} computations with pairwise distances $\sqrt{\max(\|u-v\|_2^2/D,10^{-8})}$.
At inference, full 256-cell populations are scored in raw coordinates, distances have the same numerical floor, and the final discrepancy is clamped at zero.
For the ideal energy discrepancy without the numerical floor, omitting the common dimension factor preserves ranking in exact arithmetic.

\paragraph{Mechanism experiment.}
Six independently trained configurations share common initial tensors, actual sampled batches, preprocessing, and optimization settings.
They run for the first 10,000 warmup updates with both IDR terms disabled.
Besides PED alone, they use state SIGReg across the population axis, across the joint population--cell axis, or within each population; the latter sample counts are 60, 15,360, and 256.
The remaining two configurations add alignment alone or complete PEA to population-axis SMI.
Enabled SMI and ESI weights are 1 and alignment weight is 0.25.

Diagnostics encode the actual training batch before update 0 and after each subsequent update, without predictor calls or random-state changes.
For all $N=2Bn$ observed embeddings, global scale is $\sqrt{\sum_j\|z_j-\bar z\|^2/(ND)}$.
The other measures average matched treated--control energy discrepancy and $\|\Delta z_b\|_2/\sqrt D$ over pairs.
They include population-sampling variation and are not prediction errors.
The figure shows centered 101-step moving means with truncated boundary windows, from one training seed without uncertainty bands.
Late-window means use unsmoothed updates 9,001--10,000: SMI gives global scale 0.9737, energy contrast 0.00402, and centroid contrast 0.0562; complete PEA gives 0.9697, 0.0202, and 0.1599.

\paragraph{Ablation configurations.}
The EMA tokenizer and encoder track online weights with decay increasing from 0.996 to 1 over the scheduled 200k updates.
Prediction and ranking use stop-gradient target embeddings.
Complete PEA adds a third encoder forward on treated cells through the online branch so that the displacement uses one coordinate system.
The prefix-token predictor has the same depth and width as AdaLN but uses affine RMSNorm and LayerScale initialized to 0.1; its parameter count is 16,785,920 versus 23,081,472 for AdaLN.
The Sinkhorn recipe changes both prediction and ranking discrepancies, using debiased GeomLoss with $p=2$, blur 0.50, scaling 0.9, and inputs divided by $\sqrt{512}$, while retaining auxiliary weights and ranking temperature.
Its primary checkpoint selection still uses energy-based retrieval.
Fixed-checkpoint rescoring changes only the test discrepancy.

\section{Evaluation protocol and baseline configurations}
\label{app:retrieval-protocol}
\paragraph{Queries, split, and scoring.}
\label{app:query-scoring}
The condition--background split follows PerturbDiff's Tahoe setting~\citep{yuan2026perturbdiff,zhang2026tahoe}.
Of 3,673 test drug--dose--background groups, 81 have fewer than 256 treated cells and are excluded as queries.
The remaining 3,592 queries each freeze one sample of distinct target and control cells; target sampling balances source-pool quotas, and controls reproduce the target's plate composition within the same background.
Cells are unpaired and controls can recur across queries.
The background-specific test catalogs contain 734 or 735 actions from 268 drugs, using observed test conditions; an action from an excluded query group may remain a candidate.
Positive doses are 0.05, 0.5, and $5\,\mu\mathrm M$, with two or three eligible doses per drug.
The protocol requires 2,638,682 query--candidate scores per model and includes the true source action.

Every method scores the same candidates against the observed target. Methods that condition on controls receive the same frozen control cells for every candidate. Candidate predictions use no test responses.
The primary comparison uses the same empirical energy formula in each method's native output space.
A frozen candidate-specific index breaks exact ties identically across methods.
Drug scores take the best available dose.
Dose accuracy supplies the true drug for every query: 690 queries offer two source-drug doses and 2,902 offer three.
All methods are evaluated on the complete set of 3,592 queries.

\paragraph{Aggregation and uncertainty.}
\label{app:aggregation-uncertainty}
For a per-query metric $v$, let $\mathcal C$, $\mathcal D_c$, and $\mathcal S_{cd}$ denote backgrounds, drugs with retained queries, and their retained doses.
There is one query per group, and the score is
\begin{equation}
\operatorname{Macro}(v)=\frac1{|\mathcal C|}\sum_c
\frac1{|\mathcal D_c|}\sum_d
\frac1{|\mathcal S_{cd}|}\sum_s v_{cds}.
\end{equation}
Joint Recall@$K$ uses $\mathbf1[r^{\rm J}\leq K]$, and MRR averages $1/r^{\rm J}$; drug ranks are computed after dose aggregation.
Four thousand paired bootstrap replicates resample 268 target-drug clusters with replacement, retaining linked doses and backgrounds and sharing sampled weights across compared methods.
Percentile 2.5/97.5 bounds define the 95\% intervals.
All paired differences are computed from unrounded metrics.
The intervals quantify variation across target drugs at fixed checkpoints and query cells.

\paragraph{Selection.}
\label{app:selection-protocol}
Of 270 validation groups, 263 meet the 256-treated-cell threshold and cover 19 target drugs across five backgrounds. We freeze 256 treated and 256 plate-matched control cells per query, as on test. Each background uses the same validation catalog of 379 drugs at three positive doses (1,137 candidates).
For each PerturbJEPA retrieval configuration, we use a 200,000-update training budget and select the checkpoint with the highest validation macro joint Recall@1 for test evaluation. Checkpoints are saved every 10,000 updates. For Recall@1 differences below $10^{-12}$, we break ties by joint MRR and then by earliest update. Table~\ref{tab:supp-ablation-selection} gives the selected updates and validation scores.

\paragraph{Mean and Linear.}
Mean predicts the cell-weighted training mean for each drug--dose action, pooled across training backgrounds and including implicit zero expression.
Linear fits training pseudobulk response differences using an unpenalized intercept, a background effect, and a categorical action effect, with no interaction term.
Responses subtract same-plate controls and average plates equally.
Validation response MSE selects ridge coefficient 1 from $\{0.01,0.1,1,10,100\}$ without refitting on validation.
Prediction adds the fitted response to the query-control mean.
Both use CP10K log-expression on the fixed, ordered set of 2,000 genes and return singleton distributions.

\paragraph{Released checkpoints.}
STATE~\citep{adduri2026state} uses the official ST-HVG-Tahoe few-shot checkpoint and its native transition inference path with ordered HVGs.
All 3,592 query conditions map to its declared test split, with none in its training condition list.
The two PerturbDiff releases~\citep{yuan2026perturbdiff} use the official DDIM path with start time 100, $\eta=0$, guidance 1, clipping, and float32.
Inputs are CP10K log-expression divided by 10; predictions are multiplied by 10 for scoring.
The adapter divides controls by plate, deterministically pads each slice to 256 cells with replacement, generates predictions, discards padding, and restores query order.
Initial noise is shared across candidates within a query and plate.

\paragraph{CPA.}
Official CPA~\citep{lotfollahi2023cpa} trains with its native objective on a fixed cache sampling at most 64 treated cells per plate--background--action group and 256 controls per plate--background group.
The cache has 3,720,515 treated cells, 170,948 controls, and 18,255 validation cells.
Training is configured for 20 epochs, with minimum validation negative log-likelihood selecting the checkpoint.
Native drug, continuous-dose, and background encoders are retained.
Inference decodes conditional means for the same 256 controls, without added observation noise.

\paragraph{CellFlow.}
The adaptation retains the official conditional velocity model, optimal-transport flow-matching solver, and probability path~\citep{klein2025cellflow}.
A centered non-whitened 100-component PCA fits 50,000 distinct training cells on the same ordered HVGs, retaining 45.08\% of fit-sample variance; its parameters remain frozen.
Conditioning uses a 1,024-slot radius-4 Morgan count fingerprint, $\log_{10}$ dose in nM, and a 50-dimensional background one-hot vector.
The condition encoder uses attention pooling and two 1,024-unit layers; its post-pooling output dropout is 0.9.
The time branch uses three 2,048-unit layers; state and velocity branches each use three 4,096-unit layers, with SiLU activations.
The model has 118,223,396 parameters.
Training draws 256 controls and treated cells per group, with matched backgrounds and plate-local sampling; balanced OT uses squared Euclidean cost, mean cost scaling, $\epsilon=1$, and tolerance $10^{-3}$.
The path is $z_t=(1-t)z_0+tz_1+\xi$, $\xi\sim\mathcal N(0,I)$, with target velocity $z_1-z_0$ and no condition regularization.
Adam uses learning rate $5\times10^{-5}$ and accumulation over 20 population steps; 500,000 population steps give 25,000 updates on one A100 40GB.
Evaluation uses final fixed-budget weights, without temporal averaging.
Tsit5 transports query controls from time 0 to 1 with relative/absolute tolerances $10^{-5}$, dropout disabled, and no added inference noise.
Predictions are compared in native PCA space.

Baseline comparisons assess system-level retrieval, while controlled ablations evaluate individual PerturbJEPA components.

\begin{table}[t]
\centering
\small
\setlength{\tabcolsep}{2pt}
\caption{\textbf{Additional overall joint-retrieval metrics.} All rows use the same 3,592 queries, fixed candidate catalogs, and hierarchical macro aggregation as the main comparison. Recall is a percentage and MRR is $\times100$.}
\label{tab:aggregate-extended}
\begin{tabular*}{\columnwidth}{@{\extracolsep{\fill}}lrrrr@{}}
\toprule
Model & R@1 & R@5 & R@10 & MRR\\
\midrule
Mean & 4.47 & 11.93 & 16.67 & 9.24\\
Linear & 75.57 & 92.27 & 94.98 & 82.94\\
CPA & 0.10 & 0.61 & 0.97 & 0.88\\
PerturbDiff (scratch) & 1.92 & 5.08 & 7.64 & 4.17\\
PerturbDiff (fine-tuned) & 2.93 & 7.86 & 11.20 & 5.96\\
STATE & 88.06 & 95.28 & 96.76 & 91.39\\
CellFlow & 4.63 & 16.84 & 26.42 & 11.78\\
\midrule
\textbf{PerturbJEPA} & \textbf{91.46} & \textbf{98.04} & \textbf{98.77} & \textbf{94.43}\\
\bottomrule
\end{tabular*}
\end{table}

\section{Why interventional contrast collapses and how effect alignment helps}
\label{supp:theory:contrast-theory}

We analyze PED, SMI, PEA, and ESI under the ranking-free setting of the controlled mechanism experiment in the main paper.
A response-contraction construction is followed by finite-batch results for the PEA and SIGReg losses.
Section~\ref{supp:theory:validation-30k} then reports an additional three-seed, 30k-update validation of these mechanisms.

\subsection{Scope and notation}
\label{supp:theory:theory-scope}

Let $c$ denote a matched biological and experimental context and $a$ an intervention.
Fix a population size $n$ and a sampling protocol shared by the compared conditions.
For a sampled expression population $X$, define the encoded empirical measure
\begin{equation}
 \begin{aligned}
 \widehat\mu_\theta(X)&=\frac1n\sum_{i=1}^{n}\delta_{[E_\theta(X)]_i},\\
 Q_{\theta,c}^{r}&=\operatorname{Law}\!\left(\widehat\mu_\theta(X_c^{r})\mid c,r\right),
 \qquad r\in\{0,a\}.
 \end{aligned}
 \label{supp:theory:icc-population-law}
\end{equation}
Thus $Q_{\theta,c}^{r}$ is a probability law over empirical measures.
It includes population-sampling randomness and permits each slot's embedding to depend on all cells in its own population.
The definition accommodates dependent encoded slots and unpaired control and treated populations.
Define the corresponding expression-population laws by replacing $\widehat\mu_\theta(X)$ with $n^{-1}\sum_i\delta_{x_i}$ in Equation~\eqref{supp:theory:icc-population-law}.
Exact population-level ICC at $(c,a)$ occurs when these expression-population laws differ but $Q_{\theta,c}^{a}=Q_{\theta,c}^{0}$, while the pooled state representation remains globally nondegenerate.
Global nondegeneracy separates this conditional failure from the all-constant representation.

The marginal law of a uniformly sampled encoded cell is the mean measure $\overline\mu_{\theta,c}^{r}=\mathbb E[\widehat\mu_\theta(X_c^{r})\mid c,r]$.
Equality of the full laws $Q_{\theta,c}^{a}=Q_{\theta,c}^{0}$ implies equality of these mean measures, but the converse fails.
For example, with $n=2$ and $u\ne0$, let a control set be $\{u,u\}$ or $\{-u,-u\}$ with equal probability, whereas a treated set is always $\{u,-u\}$.
Both random-slot marginals equal $(\delta_u+\delta_{-u})/2$, yet the control has zero within-set variance and the treated set does not.
Their empirical-measure laws remain distinguishable, so this is not population-level ICC under Equation~\eqref{supp:theory:icc-population-law}.

When first moments exist, equality of $Q_{\theta,c}^{a}$ and $Q_{\theta,c}^{0}$ implies zero \emph{expected} centroid contrast,
\begin{equation}
 \begin{aligned}
 \delta_\theta(c,a)={}&
 \mathbb E\!\left[\frac1n\sum_i[E_\theta(X_c^a)]_i\mid c,a\right]\\
 &-\mathbb E\!\left[\frac1n\sum_i[E_\theta(X_c^0)]_i\mid c,0\right]=0.
 \end{aligned}
 \label{supp:theory:icc-expected-contrast}
\end{equation}
The converse need not hold, and equality of these laws does not force two separately sampled populations to have identical realized centroids.
Likewise, PED compares a pair of realized empirical measures, not $Q_{\theta,c}^{a}$ and $Q_{\theta,c}^{0}$ directly.
Its finite-sample value can be positive under equal population laws.
Dependence between encoded slots does not invalidate the empirical discrepancy, but rules out importing an independent-cell null calibration without further assumptions.

For a realized batch of $B$ matched population pairs, define $\Delta x_b=\bar x_b^1-\bar x_b^0$ and $\Delta z_b=\bar z_b^1-\bar z_b^0$, where bars denote means over the $n$ cells and $z_{bi}^r=[E_\theta(X_b^r)]_i$. Let $v_b=g(\Delta x_b)$ be the learned effect target.
Let $\Delta Z,V\in\mathbb R^{B\times D}$ have rows $\Delta z_b^\top,v_b^\top$, respectively, and define
\begin{equation}
 \begin{aligned}
 R_\Delta&=\frac{\|\Delta Z\|_F}{\sqrt{BD}},&
 R_v&=\frac{\|V\|_F}{\sqrt{BD}},\\
 S_\Delta&=\frac{\|H_B\Delta Z\|_F}{\sqrt{BD}},&
 S_v&=\frac{\|H_BV\|_F}{\sqrt{BD}}.
 \end{aligned}
 \label{supp:theory:theory-scales}
\end{equation}
where $H_B=I_B-\mathbf1\mathbf1^\top/B$ centers rows.
$R$ measures root-mean-square magnitude, whereas $S$ measures variation across effects after removing their mean.
These finite-batch quantities are distinct from the main-paper diagnostic $C_{\mathrm{mean}}=B^{-1}\sum_b\|\Delta z_b\|_2/\sqrt D$, which averages norms rather than squared norms.
They describe measured contrasts; finite sampling can create nonzero measured contrasts even when population means agree.

\subsection{Objective definitions}
\label{supp:theory:objectives}

For completeness, we state the objectives used in the analysis. For finite populations $U=\{u_i\}_{i=1}^m$ and $V=\{v_j\}_{j=1}^n$, let $\rho_D(u,v)=\|u-v\|_2/\sqrt D$ and define the exact empirical energy discrepancy
\begin{equation}
\begin{split}
 \mathcal D_E(U,V)={}&\frac1{mn}\sum_{i,j}\rho_D(u_i,v_j)\\
 &-\frac1{2m^2}\sum_{i,i'}\rho_D(u_i,u_{i'})\\
 &-\frac1{2n^2}\sum_{j,j'}\rho_D(v_j,v_{j'}).
\end{split}
\label{supp:theory:energy}
\end{equation}
The V-statistic includes diagonal terms. With $Z_b^r=E_\theta(X_b^r)$ and $\widehat Z_b=P_\varphi(Z_b^0,e_\eta(a_b))$, PED is $\mathcal L_{\mathrm{PED}}=B^{-1}\sum_b\mathcal D_E(\widehat Z_b,Z_b^1)$. The encoder is shared and both state branches receive gradients. Translation and scaling identities below concern this exact discrepancy before the numerical distance floor.

The PEA objective is
\begin{equation}
 \mathcal L_{\mathrm{PEA}}(\Delta Z,V)
 =\frac1{BD}\left\|\frac{\Delta Z}{\alpha}-V\right\|_F^2,
 \qquad \alpha>0.
 \label{supp:theory:pea}
\end{equation}
The implemented value is $\alpha=0.2$. Its independently parameterized effect encoder is
\begin{equation}
 g(\Delta x)=W_2\tanh\!\left(W_1(16\Delta x)\right),
 \label{supp:theory:effectencoder}
\end{equation}
with no biases or output normalization.

For a collection $U=\{u_m\}_{m=1}^M$, let $a_k=\xi_k/\|\xi_k\|_2$ with $\xi_k\sim\mathcal N(0,I_D)$. Use $K=256$ unit directions, $L=17$ frequencies $t_\ell=\ell h$ for $\ell=0,\ldots,L-1$, and $h=3/(L-1)$. Set $\phi_\ell=e^{-t_\ell^2/2}$, $q_0=q_{L-1}=h$, and $q_\ell=2h$ otherwise. The sample-count-normalized SIGReg penalty is
\begin{equation}
\begin{aligned}
 \widehat\phi_{k\ell}(U)&=\frac1M\sum_m e^{it_\ell a_k^\top u_m},\\
 \mathcal I(U)&=\frac1K\sum_{k=1}^K\sum_{\ell=0}^{L-1}
 q_\ell\phi_\ell\left|\widehat\phi_{k\ell}(U)-\phi_\ell\right|^2.
\end{aligned}
\label{supp:theory:sigreg}
\end{equation}
The weights incorporate the symmetric negative-frequency contribution; there is no additional factor of $M$. Inputs are not centered or standardized. Directions are synchronized across workers and refreshed across calls. SMI and ESI use this same penalty with different sample collections:
\begin{equation}
\begin{aligned}
 \mathcal R_{\mathrm{SMI}}&=\frac1{2n}\sum_{r\in\{0,1\}}\sum_{i=1}^n
 \mathcal I\!\left(\{z_{bi}^r\}_{b=1}^B\right),\\
 \mathcal R_{\mathrm{ESI}}&=\mathcal I\!\left(\{v_b\}_{b=1}^B\right).
\end{aligned}
\label{supp:theory:regularization-axes}
\end{equation}
Thus state regularization is across the synchronized batch for each state and cell slot, while effect regularization is across population-level effect targets.

\subsection{A predictive objective can suppress responses without losing state dispersion}
\label{supp:theory:theory-collapse}

First consider the exact energy discrepancy in Equation~\eqref{supp:theory:energy}, before the numerical distance floor.
For any two finite populations $U,V$, common translation $h\in\mathbb R^D$, and scalar $s\geq0$,
\begin{equation}
 \mathcal D_E(h+sU,h+sV)=s\,\mathcal D_E(U,V).
 \label{supp:theory:energy-contraction}
\end{equation}
Every Euclidean distance in the cross- and within-population terms is multiplied by $s$, which proves the identity.
Consequently, if a state representation decomposes into a shared context component and a response component, contracting the response component also contracts any residual prediction discrepancy in that component.
Because the target encoder is jointly trained, this can reduce PED without improving the predictor's estimate of the original expression response.

\begin{proposition}[Response contraction with unchanged state marginals]
\label{supp:theory:icc-construction}
Consider an illustrative representation family in which each encoded population consists of $n$ repeated copies of
\begin{equation}
 z_s^r=\sqrt{1-s^2}\,H+sR^r,\qquad 0\leq s\leq1,
 \label{supp:theory:gaussian-contraction}
\end{equation}
Here $r\in\{0,1\}$, $H\sim\mathcal N(0,I_D)$ is a background shared by the matched pair, each $R^r$ has marginal law $\mathcal N(0,I_D)$, and $H$ is independent of $(R^0,R^1)$.
Suppose the predictor copies the encoded control population and ignores the action.
Then each state marginal is $\mathcal N(0,I_D)$ for every $s$, while
\begin{equation}
 \begin{aligned}
 \mathbb E\mathcal L_{\mathrm{PED}}(s)
 &=\frac{s}{\sqrt D}\,\mathbb E\|R^1-R^0\|_2,\\
 \Delta z_s&=s(R^1-R^0).
 \end{aligned}
 \label{supp:theory:contraction-loss}
\end{equation}
Thus, whenever $\mathbb P(R^1\ne R^0)>0$, decreasing $s$ strictly reduces the expected prediction loss while leaving state marginals unchanged.
At $s=0$, the prediction loss is zero and the representation contains only background.
\end{proposition}

\begin{proof}
For each $r$, the independent Gaussian terms in Equation~\eqref{supp:theory:gaussian-contraction} have covariances $(1-s^2)I_D$ and $s^2I_D$, so their sum is standard Gaussian.
All within-population distances vanish because slots are repeated.
Every cross-population distance is $\|z_s^1-z_s^0\|_2/\sqrt D=s\|R^1-R^0\|_2/\sqrt D$.
Substitution in Equation~\eqref{supp:theory:energy} gives Equation~\eqref{supp:theory:contraction-loss}.
The expectation is finite by Gaussian first moments and positive when the two residuals differ with positive probability.
\end{proof}

The construction can contain a genuine conditional intervention response despite identical unconditional state marginals.
For example, take input populations consisting of repeated observations $(H,R^r)$, with $R^0=U$ and $R^1=A$, where $U$ and a continuous intervention $A$ are independent standard Gaussians, also independent of $H$.
With context $c=H$ and action $a=A$, the control residual is random while the treated residual is fixed at $A$; the conditional input empirical-measure laws differ.
At $s=0$, both encoded populations have empirical measure $\delta_H$, so both conditional laws in Equation~\eqref{supp:theory:icc-population-law} are the same point mass at $\delta_H$.
This construction establishes a response-suppressing descent path in a representation class admitting the stated encoders and copy predictor.
The controlled training experiment in the main paper examines the corresponding contrast diagnostics in the implemented transformer.

The unchanged-marginal argument also applies to the expectation of the finite-sample SMI penalty, rather than only an ideal distribution constraint.
If batch pairs are independent, each fixed state and slot supplies $B$ independent standard Gaussian vectors.
For any unit projection and frequency $t_\ell$, the empirical characteristic function $\widehat\phi_\ell$ satisfies
\begin{equation}
 \begin{aligned}
 \mathbb E|\widehat\phi_\ell-\phi_\ell|^2
 &=\frac{1-\phi_\ell^2}{B},\\
 \mathbb E\mathcal R_{\mathrm{SMI}}
 &=\frac1B\sum_{\ell=0}^{L-1}q_\ell\phi_\ell(1-\phi_\ell^2).
 \end{aligned}
 \label{supp:theory:smi-null-expectation}
\end{equation}
The first equality follows from the variance of an average of independent complex variables of unit modulus and mean $\phi_\ell$; averaging Equation~\eqref{supp:theory:sigreg} gives the second.
The nonzero sampling baseline is independent of $s$.
Control--treated dependence does not change this calculation because their state penalties are computed separately.
SMI can therefore preserve its expected value while within-context response information disappears.
This calculation applies to the across-population SMI used by the main model.

\subsection{Alignment to a learned effect target still permits joint contraction}
\label{supp:theory:theory-alignment}

PEA introduces a matched contrast, but $v_b$ is not a fixed supervisory label.
In particular, $\Delta Z=V=0$ gives $\mathcal L_{\mathrm{PEA}}=0$ even when the observed $\Delta x_b$ are nonzero.
More generally, simultaneous contraction of the two aligned quantities gives
\begin{equation}
 \mathcal L_{\mathrm{PEA}}(s\Delta Z,sV)
 =s^2\mathcal L_{\mathrm{PEA}}(\Delta Z,V).
 \label{supp:theory:pea-contraction}
\end{equation}
This follows by factoring $s$ out of the squared residual in Equation~\eqref{supp:theory:pea}.
For an already aligned pair, the loss remains zero along the contraction; for an imperfectly aligned pair it decreases.
The effect-branch part of this contraction is directly realizable by replacing its final matrix $W_2$ with $sW_2$ in Equation~\eqref{supp:theory:effectencoder}.
The equality describes a contraction in representation-output space.
It explains why alignment alone does not fix the zero-contrast solution admitted by PED and SMI.

The output gradients show how alignment becomes informative when its effect target is retained:
\begin{equation}
 \begin{aligned}
 \frac{\partial\mathcal L_{\mathrm{PEA}}}{\partial\Delta z_b}
 &=\frac{2}{BD\alpha}\left(\frac{\Delta z_b}{\alpha}-v_b\right),\\
 \frac{\partial\mathcal L_{\mathrm{PEA}}}{\partial v_b}
 &=-\frac{2}{BD}\left(\frac{\Delta z_b}{\alpha}-v_b\right).
 \end{aligned}
 \label{supp:theory:pea-output-gradients}
\end{equation}
At $\Delta z_b=0$ and $v_b\ne0$, gradient descent on the state displacement moves it toward $\alpha v_b$.
Since $\Delta z_b=\bar z_b^1-\bar z_b^0$, the two population centroids receive opposite output gradients.
At the same time, the effect-branch gradient tends to reduce an unsupported target, so a separate effect constraint is needed.
Parameter updates additionally depend on both encoders' Jacobians.

\subsection{What the implemented effect regularizer rules out}
\label{supp:theory:theory-esi}

The following statement applies directly to the finite directions, frequencies, and positive quadrature weights of Equation~\eqref{supp:theory:sigreg}, for any effect-encoder rank.
Define the penalty of the all-zero effect batch by
\begin{equation}
 \mathcal I_0=\sum_{\ell=0}^{L-1}
 q_\ell\phi_\ell(1-\phi_\ell)^2>0.
 \label{supp:theory:esi-zero-reference}
\end{equation}
It is positive because the frequency grid includes nonzero frequencies with positive weights.

\begin{proposition}[Constant effect targets incur a nonzero ESI penalty]
\label{supp:theory:effect-isotropy}
If $v_b=v_\circ$ for every $b$, then $\mathcal I(V)\geq\mathcal I_0$, with equality at $v_\circ=0$.
Hence $\mathcal I(V)<\mathcal I_0$ certifies that the realized batch contains nonconstant effect targets.
\end{proposition}

\begin{proof}
For a constant batch, the empirical characteristic function at direction $a_k$ and frequency $t_\ell$ is $\exp(it_\ell a_k^\top v_\circ)$.
For any real $u$ and $\phi_\ell>0$,
\[
 |e^{iu}-\phi_\ell|^2
 =1+\phi_\ell^2-2\phi_\ell\cos u
 \geq(1-\phi_\ell)^2.
\]
Multiply by $q_\ell\phi_\ell$, sum over frequencies, and average over directions to obtain the claim.
At $v_\circ=0$, every term attains equality.
\end{proof}

ESI also has directions of decreasing penalty away from the all-zero effect batch.
For fixed vectors $w_b$ with $\sum_bw_b=0$, let $W$ stack the rows $w_b^\top$, set $v_b=sw_b$, and define
\[
 \begin{aligned}
 A_{\mathrm{ESI}}&=\sum_\ell q_\ell\phi_\ell(1-\phi_\ell)t_\ell^2>0,\\
 M_W&=\frac1{KB}\sum_{k,b}(a_k^\top w_b)^2.
 \end{aligned}
\]
Taylor expansion of the finite sums in Equation~\eqref{supp:theory:sigreg} gives, as $s\to0$,
\begin{equation}
 \mathcal I(sW)=\mathcal I_0-s^2A_{\mathrm{ESI}}M_W+O(s^4).
 \label{supp:theory:esi-local-expansion}
\end{equation}
Indeed, the projected cosine mean is
\[
  1-s^2t_\ell^2 B^{-1}\sum_b(a_k^\top w_b)^2/2+O(s^4).
\]
The projected sine mean is $O(s^3)$ because the first moment is zero, so its square contributes only $O(s^6)$.
Expanding the squared cosine residual yields Equation~\eqref{supp:theory:esi-local-expansion}.
For $M_W>0$, a sufficiently small nonzero $s$ lowers ESI.
ESI therefore favors effect variation along these output directions.
The full parameter update also depends on PEA, PED, and the network Jacobians.
At the exact zero configuration, both ESI and PEA have zero first derivatives; escape from that stationary point requires more than the negative curvature of ESI in output space.

\subsection{Effect variation is transferred to latent intervention contrasts}
\label{supp:theory:theory-transfer}

\begin{proposition}[Finite-batch contrast transfer]
\label{supp:theory:prop-contrast-transfer}
For any realized state and effect embeddings with $\alpha>0$, let $\epsilon$ be the direct PEA loss in Equation~\eqref{supp:theory:pea}.
Then
\begin{equation}
 \begin{aligned}
 \left|\frac{R_\Delta}{\alpha}-R_v\right|&\leq\sqrt\epsilon,\\
 \left|\frac{S_\Delta}{\alpha}-S_v\right|&\leq\sqrt\epsilon.
 \end{aligned}
 \label{supp:theory:contrast-transfer}
\end{equation}
In particular, $R_\Delta\geq\alpha(R_v-\sqrt\epsilon)_+$ and
$S_\Delta\geq\alpha(S_v-\sqrt\epsilon)_+$, where $(u)_+=\max\{u,0\}$.
\end{proposition}

\begin{proof}
By Equation~\eqref{supp:theory:pea}, $\|\Delta Z/\alpha-V\|_F=\sqrt{BD\epsilon}$.
The reverse triangle inequality for the Frobenius norm gives
$|\|\Delta Z/\alpha\|_F-\|V\|_F|\leq\sqrt{BD\epsilon}$.
Divide by $\sqrt{BD}$ to obtain the first inequality.
The matrix $H_B$ is an orthogonal projection, so
$\|H_B(\Delta Z/\alpha-V)\|_F\leq\|\Delta Z/\alpha-V\|_F$.
Apply the same reverse triangle inequality to the centered matrices to obtain the second.
Nonnegativity of the norms gives the stated one-sided forms.
\end{proof}

The two propositions identify the complementary roles of the complete PEA branch.
ESI penalizes constant effect targets, and direct PEA couples their magnitude and variation to the state-encoder displacement.
For exact alignment, $\Delta Z=\alpha V$, so $\mathcal I(V)<\mathcal I_0$ excludes a batch of identical displacements, including an all-zero displacement batch.
For approximate alignment, $S_v>\sqrt\epsilon$ suffices to guarantee positive displacement variation, and $R_v>\sqrt\epsilon$ suffices to guarantee positive displacement magnitude.
Section~\ref{supp:theory:validation-30k} measures these quantities during training.

\begin{figure*}[t]
\centering
\includegraphics[width=\textwidth]{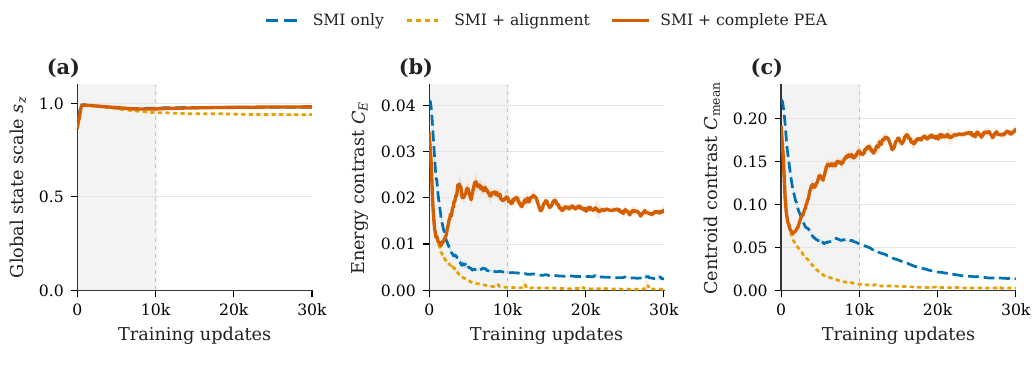}
\caption{\textbf{Additional 30k validation of global dispersion and intervention contrast.}
All variants retain PED and population-axis SMI; alignment adds the PEA loss, and complete PEA also adds ESI. Both IDR losses are disabled.
(a) Pooled, centered global state scale $s_z$.
(b) Matched observed treated--control energy contrast $C_E=B^{-1}\sum_b\mathcal D_E(Z_b^0,Z_b^1)$, using the implemented numerical distance floor.
(c) Mean dimension-normalized centroid distance $C_{\mathrm{mean}}=B^{-1}\sum_b\|\Delta z_b\|_2/\sqrt D$.
These are encoder diagnostics, not predictor errors; $C_{\mathrm{mean}}$ differs from $R_\Delta$ and $S_\Delta$ in Equation~\eqref{supp:theory:theory-scales}.
Lines and bands show the mean and sample SD across three seeds after within-seed smoothing. The gray interval marks the 10k warmup.}
\label{supp:fig:30k-contrast}
\end{figure*}

\subsection{Interpretation of the finite-batch results}
\label{supp:theory:theory-limits}

The guarantees concern measured first-order contrasts and their variation over a realized batch.
A nonzero expected centroid contrast $\delta_\theta(c,a)$ rules out equality of the corresponding population laws. Sampling can produce a nonzero realized $\Delta z_b$ even under equal laws.
Two responses with the same mean-expression shift supply the same target $g(\Delta x)$ to PEA. Differences in variance or cell-state mixtures can affect PED when the population embeddings retain them.
The batch-level constraint also permits zero effects for individual interventions and shared effects across actions.
For example, all actions within a context could share a nonzero shift $h(c)$ while these shifts vary across contexts.
Such displacements can exhibit batch-level variation and accurate PEA alignment yet carry no within-context drug identity.
IDR supplies the separate source-label objective for within-context candidate discrimination.

The effect encoder in Equation~\eqref{supp:theory:effectencoder} has a $128$-dimensional hidden layer and a $512$-dimensional output.
Thus every effect embedding lies in the image of $W_2$, a subspace of dimension at most $128$, and its covariance cannot equal $I_{512}$.
The finite projected ESI penalty regularizes the available subspace through its sampled directions.
Propositions~\ref{supp:theory:effect-isotropy} and~\ref{supp:theory:prop-contrast-transfer} remain valid under this bottleneck.

The estimated expression difference $\Delta x_b$ contains sampling variation and may contain residual experimental confounding.
The learned map $g$ can compress these differences, so biological fidelity requires evidence about which directions it retains.
The analysis establishes the geometric role of effect regularization and alignment; the retrieval experiments assess their utility for source discrimination.

\subsection{Additional 30k validation}
\label{supp:theory:validation-30k}

We extend the 10,000-update contrast diagnostics in main-paper Figure~3 to 30,000 updates for three configurations and three training seeds (42--44).

\begin{figure*}[t]
\centering
\includegraphics[width=\textwidth]{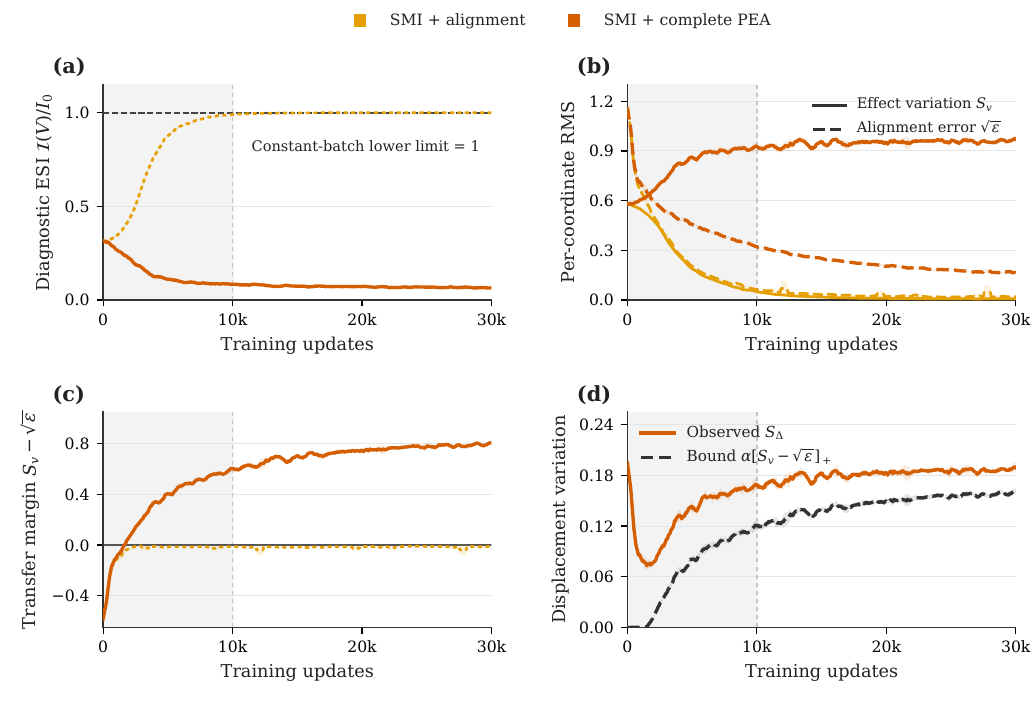}
\caption{\textbf{Effect-target variation and finite-batch transfer in the additional 30k runs.}
PED and SMI are common to both variants; complete PEA adds ESI to alignment.
(a) Diagnostic $\mathcal I(V)/\mathcal I_0$ using 256 fixed unit projections and the 17-point frequency grid of Equation~\eqref{supp:theory:sigreg}; training ESI refreshes projections across calls.
The reference line is the constant-batch lower limit; a raw value below 1 certifies nonconstant targets by Proposition~\ref{supp:theory:effect-isotropy}.
(b) Target variation $S_v$ (solid) and unweighted alignment error $\sqrt\epsilon$ (dashed), with color identifying the variant.
(c) Signed transfer margin $m_S=S_v-\sqrt\epsilon$.
(d) FULL's measured $S_\Delta$ and lower bound $L_S=0.2[m_S]_+$ from Proposition~\ref{supp:theory:prop-contrast-transfer}.
SB has no effect branch and is omitted.
Quantities are computed per batch before the smoothing and three-seed aggregation used in Figure~\ref{supp:fig:30k-contrast}.}
\label{supp:fig:30k-mechanism}
\end{figure*}

\paragraph{Training and statistical protocol.}
All three configurations retain PED and population-axis SMI, with both IDR losses disabled.
Here SB denotes SMI only, ALIGN adds direct PEA alignment, and FULL adds both alignment and ESI.
Enabled SMI and ESI weights are 1, the alignment weight is 0.25, and $\alpha=0.2$.
Within each seed, the configurations share initial tensors and actual sampled training batches.
Four workers supply $B=60$ matched population pairs per update, with 256 cells per population and $D=512$.
We retain the 200k learning-rate schedule, including 10k warmup updates, and stop each run at 30k.

Diagnostics are measured before update 0 and after every 50 updates, giving 601 records per run.
Curves first apply an 11-observation centered moving mean within each seed (a 500-update span, shortened at boundaries), then show the across-seed mean and sample standard deviation (SD).
Unless stated otherwise, late-window values below are means of the three within-seed medians of the 20 raw observations in updates 29,001--30,000.
Margins, bounds and normalized quantities are computed per raw batch before any aggregation.

\paragraph{Global dispersion and measured contrasts.}
All three variants retain similar global scale in Figure~\ref{supp:fig:30k-contrast}. Late-window $s_z$ is 0.97895 for SMI only, 0.93905 for alignment, and 0.98042 for complete PEA.
Complete PEA gives energy contrast 0.01673 and centroid contrast 0.18495, compared with 0.002484 and 0.01348 for SMI only, and 0.0002465 and 0.002354 for alignment.
Thus sustained global dispersion is compatible with much smaller treatment--control contrasts, while adding ESI to alignment retains larger measured responses.

\begin{figure*}[t]
\centering
\includegraphics[width=\textwidth]{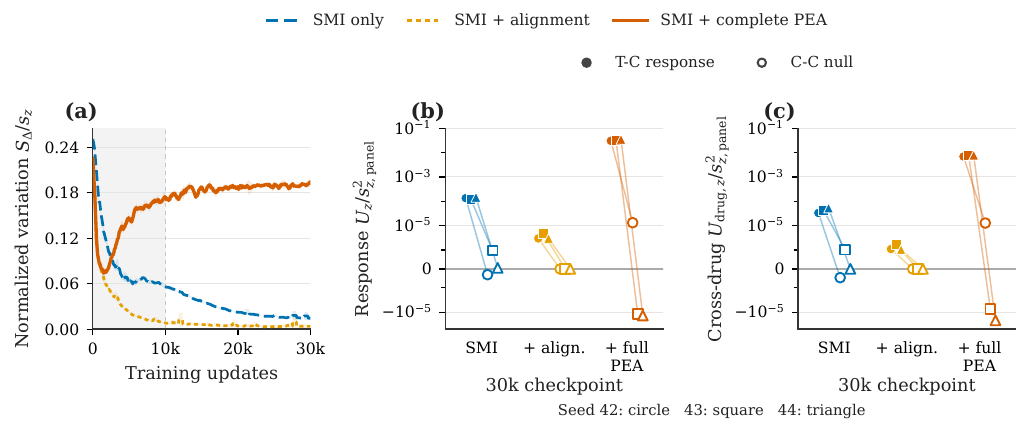}
\caption{\textbf{Additional checks for scale, sampling variation and shared context shifts.}
All three training variants retain PED and population-axis SMI with both IDR terms disabled.
(a) Training-batch variation $S_\Delta/s_z$, with the same three-seed mean, SD and smoothing as Figure~\ref{supp:fig:30k-contrast}.
(b) Cross-draw response $U_z/s_{z,\mathrm{panel}}^2$ from Equation~\eqref{supp:theory:cross-draw-response} and its matched control--control null at 30k.
(c) Within-context cross-drug variation $U_{\mathrm{drug},z}/s_{z,\mathrm{panel}}^2$ from Equation~\eqref{supp:theory:cross-drug-response}, with its corresponding null.
Filled and hollow markers show response and null; each connecting line pairs one training seed, with circles, squares and triangles identifying seeds 42, 43 and 44.
Panels (b,c) use identical symmetric-log limits, linear for absolute values at most $10^{-6}$; negative estimates are retained.
The frozen panel comprises 60 conditions, five cell lines, 17 eligible contexts and 20 draws per condition.}
\label{supp:fig:30k-response}
\end{figure*}

\paragraph{When the transfer bound is positive.}
Figure~\ref{supp:fig:30k-mechanism} directly measures the quantities in Propositions~\ref{supp:theory:effect-isotropy} and~\ref{supp:theory:prop-contrast-transfer}.
The fixed-projection diagnostic ESI uses $\mathcal I_0=0.4020475797$.
Its late-window ratio is 0.999805 for ALIGN and 0.06524 for FULL.
ALIGN's late-window penalty is close to the constant-batch threshold. Its target variation ($S_v=0.00712$) is smaller than its alignment error ($\sqrt\epsilon=0.01729$), where $\epsilon$ is the unweighted loss in Equation~\eqref{supp:theory:pea}.
FULL retains $S_v=0.96587$ with $\sqrt\epsilon=0.16309$.
The corresponding late-window margins, computed before summarizing, are $-0.00998$ and $0.79352$.

In each FULL seed, all 20 late-window batches have a positive lower bound; no ALIGN batch in that window does.
FULL's observed $S_\Delta$ is $0.18702\pm0.00298$ and its lower bound $L_S$ is $0.15870\pm0.00406$, where $\pm$ denotes the sample SD of the three seed medians.
The mean of the within-seed median ratios $L_S/S_\Delta$ is $84.95\%$ (SD 0.31 percentage points).

\paragraph{Finite-pool response checks.}
Figure~\ref{supp:fig:30k-response}(a) divides each batch's $S_\Delta$ by its contemporaneous $s_z$.
The late-window ratios are 0.01414, 0.003394 and 0.19079 for SB, ALIGN and FULL, respectively, so a uniform rescaling of all state embeddings cannot explain the difference.
At the prespecified 30k checkpoint, we evaluate the encoders on a frozen panel of 60 validation treatment conditions from five cell lines.
For each condition, $R=20$ draws sample 256 cells per population from the observed finite pools and re-encode complete populations.
If $d_r$ is the treated-minus-control centroid difference in draw $r$, define the additional diagnostic
\begin{equation}
 U_z=\frac{1}{R(R-1)D}\sum_{r\ne s}d_r^\top d_s.
 \label{supp:theory:cross-draw-response}
\end{equation}
Under independent and identically distributed draws conditional on the pools, $U_z$ estimates $\|\mathbb E[d_r\mid\mathrm{pools}]\|_2^2/D$ without the squared sampling-error term in a single-draw norm.
The matched null applies the same estimator to two independently drawn control populations.

To remove a shift shared by drugs within a context, group conditions by plate, cell line and dose.
For each eligible context with $J\geq3$ drugs and one condition per drug, let
\begin{equation}
 \begin{aligned}
 \widetilde d_{jr}&=d_{jr}-\frac1J\sum_{k=1}^Jd_{kr},\\
 U_{\mathrm{drug},z}&=\frac{1}{JR(R-1)D}
 \sum_{j=1}^J\sum_{r\ne s}\widetilde d_{jr}^{\top}\widetilde d_{js}.
 \end{aligned}
 \label{supp:theory:cross-drug-response}
\end{equation}
There are 17 eligible contexts.
The null undergoes the same within-context centering.
Condition summaries weight cell lines equally, then drugs, then conditions; context summaries weight cell lines equally, then contexts.
Both statistics are divided by the checkpoint's pooled true T/C panel variance $s_{z,\mathrm{panel}}^2$, using the same denominator for response and null.
This denominator differs from the training-batch scale in panel (a); the normalized statistics are descriptive ratios rather than unbiased estimates.

At 30k, the three-seed means of normalized $U_z$ are $1.344\times10^{-4}$, $3.603\times10^{-6}$ and $3.274\times10^{-2}$ for SB, ALIGN and FULL.
The corresponding within-context cross-drug values are $4.296\times10^{-5}$, $1.262\times10^{-6}$ and $7.543\times10^{-3}$.
FULL's matched null means are $-4.240\times10^{-6}$ and $-5.491\times10^{-6}$, respectively; finite-sample cross-draw estimates may be negative and are not clipped or square-rooted.
Complete PEA retains mean-response variation across draws above the matched null, including after centering by plate, cell line and dose. These estimates condition on the observed pools, where cells may recur across draws and matched controls may overlap training.

\paragraph{Diagnostic reproducibility.}
Both finite-batch bounds held at every recorded step for ALIGN and FULL, the two variants with effect targets. In a separate 100-update comparison with matched training cells and random-generator states, enabling diagnostics produced a maximum parameter difference of $3.8091\times10^{-7}$ and failed the prespecified equality check (absolute tolerance $10^{-7}$, relative tolerance $10^{-6}$). Diagnostics were enabled in all nine reported runs.

\newpage

\section{Retrieval ablation settings and complete results}
\label{app:ablation-details}

\subsection{Objectives and comparison design}
\label{app:ablation-objectives}

The retrieval ablations comprise thirteen configurations using 2,000 highly variable genes (HVG2000) and one configuration using all 62,710 available genes.
Each configuration is trained independently from scratch.
The tokenizer, conditioning, and discrepancy comparisons use the five-term drug-IDR reference. The learned alignment heads use the full six-term model as their reference.

For the shared-encoder configurations, define
\begin{equation}
\begin{aligned}
 \mathcal L_{\mathrm{shared}}&=\mathcal L_{\mathrm{PED}}+\mathcal R_{\mathrm{SMI}},\\
 \mathcal L_{\mathrm{complete}}&=\mathcal L_{\mathrm{shared}}
 +0.25\mathcal L_{\mathrm{PEA}}+\mathcal R_{\mathrm{ESI}}.
\end{aligned}
\label{eq:supp-ablation-base}
\end{equation}
Here PED matches predicted and observed populations, SMI regularizes state marginals, PEA aligns pseudobulk effects with latent displacements, and ESI regularizes the learned effect targets.
Complete PEA includes both the alignment loss and ESI.

\paragraph{Effect preservation with drug ranking fixed.}
The no-effect configuration optimizes $\mathcal L_{\mathrm{shared}}+10^{-3}\mathcal L_{\mathrm{IDR}}^{\mathrm{drug}}$.
The alignment-only configuration adds $0.25\mathcal L_{\mathrm{PEA}}$.
Adding $\mathcal R_{\mathrm{ESI}}$ gives the reference objective $\mathcal L_{\mathrm{complete}}+10^{-3}\mathcal L_{\mathrm{IDR}}^{\mathrm{drug}}$.

\paragraph{Ranking with complete PEA fixed.}
The no-IDR configuration optimizes $\mathcal L_{\mathrm{complete}}$ while retaining both drug and numerical-dose encoders and the intervention-conditioned predictor.
Adding $10^{-3}\mathcal L_{\mathrm{IDR}}^{\mathrm{drug}}$ recovers the same reference used above.
The full model further adds $10^{-3}\mathcal L_{\mathrm{IDR}}^{\mathrm{dose}}$.
Candidate mining, subset sizes, temperature, gradient paths, and the separate drug/dose softmax normalizers are unchanged from Section~\ref{app:training-ranking}.

\paragraph{Regularization and alignment scale.}
The shared-encoder configurations apply sample-count-normalized SMI separately to each state and cell slot across population pairs, as specified in Section~\ref{app:regularizer}.
ESI acts on population-level effect embeddings.
Direct PEA uses the alignment loss
\begin{equation}
 \mathcal L_{\mathrm{PEA}}=
 \frac{1}{BD\alpha^2}\sum_{b=1}^{B}
 \|\Delta z_b-\alpha g(\Delta x_b)\|_2^2,
 \qquad \alpha=0.2.
 \label{eq:supp-ablation-alignment}
\end{equation}
Here $B$ is the synchronized population-pair count, $D=512$, and $\Delta z_b$ and $\Delta x_b$ are the treated--control centroid and expression differences.
The factor $\alpha^{-2}=25$ retains the alignment loss scale when it is written in displacement units.
The encoder architecture and input preprocessing follow Section~\ref{app:input-architecture}.

\paragraph{Learned effect alignment.}
Let $\delta_b=\Delta z_b/\alpha$ and $v_b=g(\Delta x_b)$. The direct loss above compares $\delta_b$ with $v_b$. We replace $\delta_b$ in this loss with either $h_E([\bar z_b^0;\bar z_b^1])$, using the concatenated observed control and treated centroids, or $h_\Delta(\delta_b)$, using their scaled difference.
Each head has two 512-unit hidden layers with LayerNorm and GELU and a linear 512-dimensional output. Gradients update the head, both state branches, and $g$. All six loss weights and the test-time energy scorer are the same as in the full model.
The paired joint R@1 gains for direct alignment are $9.74$ points over $h_E$ (95\% CI $[7.98,11.43]$) and $33.29$ over $h_\Delta$ ($[31.00,35.57]$).

\subsection{Controlled changes to the model and discrepancy}
\label{app:ablation-implementations}

\paragraph{Momentum-target JEPA.}
The two EMA configurations replace the shared target path with separate online and target encoders and omit SMI.
Both optimize PED with drug and dose IDR; the effect-preserving configuration additionally includes $0.25\mathcal L_{\mathrm{PEA}}+\mathcal R_{\mathrm{ESI}}$.
PED and both IDR terms compare predictor outputs with stop-gradient EMA target embeddings.
For PEA, treated populations are also encoded online, so their displacement from online controls is measured in a common coordinate system.
PEA gradients update both online state branches and the expression-effect encoder; ESI regularizes the effect embeddings.
The EMA decay increases from $0.996$ to $1$ over the scheduled 200,000 updates.
Both EMA configurations use four workers with 15 population pairs per worker, or 60 pairs per update, whereas the shared-encoder suite uses 240 pairs.
Adding complete PEA increases the population-encoder forward passes from two to three.
The within-EMA comparison keeps the target architecture and batch size fixed; comparisons with the shared-encoder family additionally change target updates, batch size, and regularization.

\paragraph{Expression-weight normalization.}
For a nonzero expression vector $x$, the three cell-token constructions are
\begin{equation}
\begin{aligned}
 t_{\ell_2}(x)&=\frac{\sum_gx_gE_g}{\sqrt{\sum_gx_g^2}},\qquad
 t_{\ell_1}(x)=\frac{\sum_gx_gE_g}{\sum_g|x_g|},\\
 t_{\mathrm{none}}(x)&=\sum_gx_gE_g.
\end{aligned}
\label{eq:supp-ablation-token-sum}
\end{equation}
The reference uses $\ell_2$ normalization; the unnormalized alternative removes only its denominator.
Because expression is nonnegative, the $\ell_1$ rule forms an expression-weighted mean of gene embeddings.
These choices change expression weights while retaining gene embeddings and population attention; they do not normalize the resulting token vector.
Normalized modes use the learned empty-cell token when their denominator is nonpositive.
The unnormalized mode uses that token when the sparse expression input contains no entries.

\paragraph{Intervention conditioning.}
The prefix-token alternative retains the $\ell_2$ tokenizer and drug-IDR objective, but passes $[u_d;u_s;Z^0]$ through a four-layer set transformer with full self-attention.
Only the cell-token outputs form the predicted population; the two intervention-token outputs are discarded.
Both predictors have width 512, eight attention heads, feed-forward width 2,048, and the same drug and dose encoders.
The prefix-token blocks use affine RMSNorm and residual LayerScale initialized to $0.1$, whereas the AdaLN reference uses zero-initialized adaptive modulation.
The prefix predictor contains 16,785,920 parameters, compared with 23,081,472 for AdaLN.

\paragraph{Population-matching discrepancy.}
The reference uses energy discrepancy in both PED and the drug-IDR candidate scores.
The alternative uses debiased GeomLoss Sinkhorn divergence~\citep{feydy2019sinkhorn} with $p=2$, blur $0.50$, scaling $0.9$, no unbalanced-mass relaxation, and the tensorized backend.
Both input populations are converted to float32 and divided by $\sqrt{512}$ before either training discrepancy is evaluated.
Writing the resulting divergence as $\mathcal D_S$, the Sinkhorn objective is
\begin{equation}
\begin{aligned}
 \mathcal L_S={}&\frac1B\sum_b\mathcal D_S(\widehat Z_b(a_b),Z_b^1)
 +\mathcal R_{\mathrm{SMI}}\\
 &+0.25\mathcal L_{\mathrm{PEA}}+\mathcal R_{\mathrm{ESI}}
 +10^{-3}\mathcal L_{\mathrm{IDR},S}^{\mathrm{drug}}.
\end{aligned}
\label{eq:supp-ablation-sinkhorn}
\end{equation}
The subscript $S$ denotes using Sinkhorn candidate distances in the drug-ranking cross-entropy.
Ranking temperature, candidate mining, subset size, and auxiliary weights are retained.
The substitution changes the form and scale of both predictive and ranking supervision.
Checkpoint selection and the primary comparison use the common energy-based retrieval scorer for both models.
Section~\ref{app:distance-sensitivity} changes only the test scorer at these already selected checkpoints.

\subsection{Training, validation selection, and test uncertainty}
\label{app:ablation-selection}

Table~\ref{tab:supp-ablation-selection} reports selection on the common validation panel from Section~\ref{app:selection-protocol}. The full model uses the same checkpoint in the overall comparison and the ablation.
Table~\ref{tab:supp-ablation-confidence} gives marginal 95\% intervals on the common 3,592 test queries using the target-drug bootstrap from Section~\ref{app:aggregation-uncertainty}.

\begin{table*}[t]
\centering
\small
\setlength{\tabcolsep}{4pt}
\caption{\textbf{Selected checkpoints and validation performance.} Each configuration maximizes hierarchical macro joint Recall@1 on the common validation panel, with joint MRR and then the earliest update breaking ties. Recall is a percentage; MRR is multiplied by $100$.}
\label{tab:supp-ablation-selection}
\begin{tabular*}{\textwidth}{@{\extracolsep{\fill}}lrrr@{}}
\toprule
Configuration & \shortstack{Selected\\(k updates)} & \shortstack{Val. joint\\R@1} & \shortstack{Val. joint\\MRR}\\
\midrule
Complete PEA, no IDR & 180 & 77.89 & 85.35\\
Drug IDR, no effect & 10 & 0.00 & 0.48\\
Drug IDR, alignment only & 20 & 12.46 & 22.41\\
Complete PEA + drug IDR & 120 & 86.67 & 91.45\\
Full HVG2000 model & 160 & 91.58 & 94.92\\
Learned endpoints & 70 & 79.65 & 87.07\\
Learned displacement & 160 & 59.12 & 70.82\\
EMA, no effect & 10 & 4.91 & 10.60\\
EMA + complete PEA & 90 & 75.44 & 82.97\\
Unnormalized weights & 180 & 87.54 & 91.07\\
$\ell_1$ weights & 30 & 35.09 & 49.48\\
Intervention prefix tokens & 140 & 72.63 & 80.55\\
Sinkhorn recipe & 30 & 61.40 & 72.22\\
\midrule
Full model, all genes & 120 & 98.60 & 99.01\\
\bottomrule
\end{tabular*}
\end{table*}

\begin{table*}[t]
\centering
\small
\setlength{\tabcolsep}{3pt}
\caption{\textbf{95\% target-drug cluster bootstrap intervals for the selected checkpoints.} Joint Recall@1 and dose accuracy use percentage units; MRR is $\times100$. These intervals condition on the trained models and frozen query cells.}
\label{tab:supp-ablation-confidence}
\begin{tabular*}{\textwidth}{@{\extracolsep{\fill}}lrrr@{}}
\toprule
Configuration & Joint R@1 & Joint MRR & Dose Acc.\\
\midrule
Complete PEA, no IDR & [80.65, 84.09] & [87.15, 89.69] & [95.10, 96.74]\\
Drug IDR, no effect & [0.45, 1.69] & [1.61, 3.20] & [36.81, 38.96]\\
Drug IDR, alignment only & [16.86, 20.30] & [28.06, 31.51] & [82.49, 85.81]\\
Complete PEA + drug IDR & [85.86, 88.92] & [90.97, 93.09] & [95.69, 97.18]\\
Full HVG2000 model & [90.01, 92.82] & [93.33, 95.43] & [98.98, 99.60]\\
Learned endpoints & [79.64, 83.77] & [86.21, 89.40] & [97.44, 98.92]\\
Learned displacement & [55.73, 60.61] & [68.84, 72.75] & [94.05, 95.97]\\
EMA, no effect & [3.53, 5.59] & [8.66, 11.18] & [52.91, 57.08]\\
EMA + complete PEA & [74.93, 78.96] & [82.73, 85.72] & [94.71, 96.30]\\
Unnormalized weights & [85.84, 89.11] & [90.78, 93.10] & [95.68, 97.12]\\
$\ell_1$ weights & [36.40, 41.02] & [50.22, 54.37] & [85.37, 88.45]\\
Intervention prefix tokens & [71.44, 75.68] & [80.03, 83.43] & [91.25, 93.63]\\
Sinkhorn recipe & [60.36, 65.36] & [71.29, 75.47] & [90.23, 92.58]\\
\midrule
Full model, all genes & [93.27, 95.59] & [95.81, 97.58] & [97.06, 98.31]\\
\bottomrule
\end{tabular*}
\end{table*}

\subsection{Full input-gene-universe comparison}
\label{app:gene-coverage}

\paragraph{Matched recipe and input dimensions.}
The complete HVG2000 and all-gene models use the same six loss weights, model width 512, encoder depth eight, predictor depth four, $\ell_2$ pooling, and training settings from Section~\ref{app:training-ranking}.
Both use 16 workers and 15 population pairs per worker, giving 240 pairs per update.
The input gene universe expands from 2,000 HVGs to all 62,710 genes.
Both use the full-library CP10K log-expression preprocessing in Section~\ref{app:input-architecture}; the all-gene model omits HVG selection.
The effect-input projection grows from $128\times2{,}000$ to $128\times62{,}710$, and the gene-embedding vocabulary expands to the same input universe.
These matrices increase parameter count, so the comparison varies gene coverage and input-layer capacity together while keeping hidden dimensions and layer counts fixed.
Prediction and retrieval continue to operate in embedding space.

\begin{table*}[t]
\centering
\small
\setlength{\tabcolsep}{3pt}
\caption{\textbf{Paired full-gene versus HVG2000 comparison.} Both models use the complete six-term objective and their independently validation-selected checkpoints. Differences are full-gene minus HVG2000, computed before rounding. Intervals are paired 95\% target-drug cluster bootstrap intervals. Recall and accuracy use percentage units; MRR is multiplied by $100$. Both dose metrics supply the true drug.}
\label{tab:supp-gene-coverage-paired}
\begin{tabular*}{\textwidth}{@{\extracolsep{\fill}}lrrrr@{}}
\toprule
Metric & HVG2000 & All genes & Difference & Paired 95\% interval\\
\midrule
Joint Recall@1 & 91.46 & 94.50 & $+3.04$ & $[1.57,4.56]$\\
Joint MRR & 94.43 & 96.81 & $+2.38$ & $[1.38,3.42]$\\
Drug Recall@1 & 91.74 & 96.30 & $+4.56$ & $[3.16,6.09]$\\
Conditional dose accuracy & 99.31 & 97.72 & $-1.60$ & $[-2.31,-0.92]$\\
Log-dose MAE $\downarrow$ & 0.0093 & 0.0300 & $+0.0207$ & $[0.0109,0.0311]$\\
\bottomrule
\end{tabular*}
\end{table*}

\paragraph{Retrieval gains and dose trade-off.}
Table~\ref{tab:supp-gene-coverage-paired} reports all paired differences and intervals from this comparison.
All-gene input improves joint Recall@1 by 3.04 percentage points, joint MRR by 2.38 units, and drug Recall@1 by 4.56 points, while reducing conditional dose accuracy by 1.60 points.
Mean absolute $\log_{10}$ dose error increases by $0.0207$, with paired interval $[0.0109,0.0311]$.
The two models reach nearly identical joint Recall@50 ($99.69\%$ for all genes and $99.71\%$ for HVG2000).
The results support improved source-drug recovery and front-of-list joint retrieval for this configuration, with weaker conditional dose discrimination.

\subsection{Complete aggregate retrieval metrics}
\label{app:ablation-full-metrics}

Tables~\ref{tab:supp-ablation-joint} and~\ref{tab:supp-ablation-drug-dose} report all 14 metrics for each configuration.
For source-action rank $r_q^{\mathrm J}$ among $C_q$ candidates, the additional per-query measures are
\begin{equation}
\begin{aligned}
 \mathrm{nDCG@10}_q&=\frac{\mathbf1[r_q^{\mathrm J}\leq10]}{\log_2(1+r_q^{\mathrm J})},\\
 \mathrm{NR}_q&=\frac{r_q^{\mathrm J}-1}{C_q-1},\\
 \mathrm{DoseErr}_q&=|\log_{10}\widehat s_q-\log_{10}s_q^\star|.
\end{aligned}
\label{eq:supp-ablation-extended-metrics}
\end{equation}
There is one relevant source action per query, so nDCG has unit ideal gain.
Drug MRR averages reciprocal source-drug rank after best-dose aggregation.
Dose error supplies the true drug for every query and selects its best-scoring positive dose, including queries with an incorrect highest-ranked drug.
All metrics follow the same hierarchical macro weighting; normalized rank and dose error are lower-is-better.
Recall and accuracy are percentages, MRR is multiplied by 100, and nDCG remains on $[0,1]$.

The extended metrics locate the ranking gains near the front of the candidate list.
Adding both IDR terms to complete PEA raises joint Recall@1 from $82.45\%$ to $91.46\%$, whereas Recall@50 changes from $99.15\%$ to $99.71\%$.
Thus ranking supervision improves the position of the source intervention, which the no-IDR model often already places among its first 50 predictions.

\begin{table*}[t]
\centering
\small
\setlength{\tabcolsep}{3pt}
\caption{\textbf{Complete joint retrieval metrics.} Recall values are percentages, MRR is $\times100$, and nDCG@10 is on $[0,1]$. Normalized rank (NR) is lower-is-better; all other columns are higher-is-better. Definitions follow Appendix~\ref{app:ablation-full-metrics}.}
\label{tab:supp-ablation-joint}
\begin{tabular*}{\textwidth}{@{\extracolsep{\fill}}lrrrrrrrr@{}}
\toprule
Configuration & R@1 & R@5 & R@10 & R@20 & R@50 & Joint MRR & nDCG@10 & NR $\downarrow$\\
\midrule
Complete PEA, no IDR & 82.45 & 95.80 & 97.78 & 98.63 & 99.15 & 88.49 & 0.9073 & 0.002452\\
Drug IDR, no effect & 1.02 & 2.55 & 3.56 & 5.53 & 10.74 & 2.36 & 0.0213 & 0.463000\\
Drug IDR, alignment only & 18.57 & 41.53 & 51.54 & 62.62 & 77.99 & 29.79 & 0.3383 & 0.055814\\
Complete PEA + drug IDR & 87.47 & 97.32 & 98.43 & 98.84 & 99.44 & 92.07 & 0.9362 & 0.001660\\
Full HVG2000 model & 91.46 & 98.04 & 98.77 & 99.13 & 99.71 & 94.43 & 0.9547 & 0.000965\\
Learned endpoints & 81.73 & 95.76 & 97.37 & 98.47 & 99.10 & 87.84 & 0.9012 & 0.003120\\
Learned displacement & 58.18 & 86.98 & 93.27 & 96.75 & 98.90 & 70.76 & 0.7600 & 0.004716\\
EMA, no effect & 4.53 & 13.59 & 19.90 & 27.82 & 42.65 & 9.90 & 0.1115 & 0.186857\\
EMA + complete PEA & 77.00 & 93.73 & 96.89 & 98.11 & 99.25 & 84.26 & 0.8725 & 0.002501\\
Unnormalized weights & 87.55 & 97.44 & 98.50 & 99.03 & 99.31 & 91.99 & 0.9358 & 0.001980\\
$\ell_1$ weights & 38.72 & 68.73 & 79.56 & 88.17 & 94.87 & 52.29 & 0.5814 & 0.013962\\
Intervention prefix tokens & 73.64 & 91.84 & 95.17 & 96.82 & 98.37 & 81.80 & 0.8495 & 0.005075\\
Sinkhorn recipe & 62.89 & 86.79 & 92.31 & 95.58 & 97.70 & 73.39 & 0.7776 & 0.008991\\
\midrule
Full model, all genes & 94.50 & 99.30 & 99.44 & 99.49 & 99.69 & 96.81 & 0.9748 & 0.001114\\
\bottomrule
\end{tabular*}
\end{table*}

\begin{table*}[t]
\centering
\small
\setlength{\tabcolsep}{3pt}
\caption{\textbf{Complete drug and conditional-dose metrics.} Recall and accuracy are percentages; MRR is $\times100$. Dose error is mean absolute error in $\log_{10}$ dose and is lower-is-better. Both dose metrics supply the true drug.}
\label{tab:supp-ablation-drug-dose}
\begin{tabular*}{\textwidth}{@{\extracolsep{\fill}}lrrrrrr@{}}
\toprule
Configuration & Drug R@1 & Drug R@5 & Drug R@10 & Drug MRR & Dose Acc. & Log-dose error $\downarrow$\\
\midrule
Complete PEA, no IDR & 85.12 & 96.72 & 98.42 & 90.28 & 95.97 & 0.0540\\
Drug IDR, no effect & 1.92 & 4.41 & 6.61 & 4.22 & 37.89 & 0.8613\\
Drug IDR, alignment only & 20.72 & 44.90 & 56.26 & 32.75 & 84.18 & 0.2038\\
Complete PEA + drug IDR & 89.83 & 97.98 & 98.94 & 93.57 & 96.47 & 0.0495\\
Full HVG2000 model & 91.74 & 98.24 & 98.92 & 94.65 & 99.31 & 0.0093\\
Learned endpoints & 82.23 & 96.21 & 97.79 & 88.35 & 98.24 & 0.0228\\
Learned displacement & 60.18 & 88.66 & 94.62 & 72.68 & 95.05 & 0.0696\\
EMA, no effect & 7.33 & 20.80 & 30.26 & 15.14 & 55.00 & 0.5666\\
EMA + complete PEA & 79.55 & 95.15 & 97.86 & 86.28 & 95.55 & 0.0587\\
Unnormalized weights & 89.99 & 97.98 & 98.94 & 93.56 & 96.43 & 0.0475\\
$\ell_1$ weights & 42.47 & 73.40 & 84.37 & 56.45 & 86.96 & 0.1670\\
Intervention prefix tokens & 78.17 & 93.61 & 96.60 & 85.09 & 92.50 & 0.1030\\
Sinkhorn recipe & 66.72 & 89.61 & 94.10 & 76.61 & 91.47 & 0.1120\\
\midrule
Full model, all genes & 96.30 & 99.44 & 99.61 & 97.84 & 97.72 & 0.0300\\
\bottomrule
\end{tabular*}
\end{table*}

\subsection{Retrieval-distance sensitivity at fixed checkpoints}
\label{app:distance-sensitivity}

\paragraph{Comparison design.}
The $2\times2$ experiment crosses energy or Sinkhorn training with energy or Sinkhorn test scoring.
Both training recipes retain SMI, complete PEA, and drug IDR with dose IDR disabled. For each recipe, we select a checkpoint using energy scores on the validation panel and evaluate both test scorers at that checkpoint.

\paragraph{Fixed scoring profiles.}
The energy profile is the raw-coordinate retrieval scorer in Section~\ref{app:training-ranking}.
The Sinkhorn profile uses the training divergence and coordinate scaling in Section~\ref{app:ablation-implementations}, with float32 arithmetic, autocast disabled, and final divergences clamped at zero.
Both profiles evaluate full 256-cell predicted and target populations with identical query cells, candidate inputs, model evaluation mode, and tie-breaking rules.
All Sinkhorn scores are finite and positive, with no top-one ties.

\paragraph{Paired comparisons.}
The bootstrap uses identical sampled drug weights across all four evaluations.
Table~\ref{tab:supp-distance-paired} separates a change of scorer at one checkpoint from comparisons between training recipes under a common scorer.
At the energy-trained checkpoint, Sinkhorn scoring changes joint Recall@1 by $-0.72$ points (95\% interval $[-1.16,-0.31]$) and MRR by $-0.56$ ($[-0.83,-0.31]$).
At the Sinkhorn-trained checkpoint, the corresponding changes are $+0.18$ ($[-0.17,0.54]$) and $+0.11$ ($[-0.09,0.31]$).
Under Sinkhorn scoring, the Sinkhorn-trained minus energy-trained difference remains $-23.67$ Recall@1 points ($[-26.04,-21.30]$) and $-18.01$ MRR units ($[-19.93,-16.21]$).
Matching training and test discrepancies therefore does not close the retrieval gap at these checkpoints.
Table~\ref{tab:supp-distance-metrics} reports all 14 metrics under all four training/scoring combinations.

\begin{table*}[t]
\centering
\small
\setlength{\tabcolsep}{3pt}
\caption{\textbf{Paired contrasts of test scorers and fixed checkpoints.} Differences and 95\% intervals use the same 4,000 target-drug bootstrap replicates across all four training/scoring combinations. Recall differences are percentage points. Each training recipe uses its independently validation-selected checkpoint, held fixed when changing the test scorer.}
\label{tab:supp-distance-paired}
\begin{tabular*}{\textwidth}{@{\extracolsep{\fill}}lrr@{}}
\toprule
Comparison & $\Delta$ Joint R@1 & $\Delta$ Joint MRR\\
\midrule
\multicolumn{3}{l}{\textit{Change test scorer: Sinkhorn minus energy, checkpoint fixed}}\\
Energy-trained & -0.72 [-1.16, -0.31] & -0.56 [-0.83, -0.31]\\
Sinkhorn-trained & +0.18 [-0.17, 0.54] & +0.11 [-0.09, 0.31]\\
\midrule
\multicolumn{3}{l}{\textit{Compare checkpoints: Sinkhorn-trained minus energy-trained}}\\
Energy scoring & -24.57 [-26.89, -22.31] & -18.68 [-20.61, -16.87]\\
Sinkhorn scoring & -23.67 [-26.04, -21.30] & -18.01 [-19.93, -16.21]\\
\bottomrule
\end{tabular*}
\end{table*}

\begin{table}[t]
\centering
\small
\setlength{\tabcolsep}{2pt}
\caption{\textbf{Complete metrics for retrieval-distance sensitivity.} E and S denote energy and Sinkhorn, respectively; columns specify training/test discrepancy. Recall and dose accuracy are percentages, MRR is $\times100$, and nDCG remains on $[0,1]$. Dose metrics supply the true drug. Each training recipe uses one fixed checkpoint for both test scorers.}
\label{tab:supp-distance-metrics}
\begin{tabular*}{\columnwidth}{@{\extracolsep{\fill}}lrrrr@{}}
\toprule
Metric & E/E & E/S & S/E & S/S\\
\midrule
Joint R@1 & 87.47 & 86.74 & 62.89 & 63.07\\
Joint R@5 & 97.32 & 97.06 & 86.79 & 87.10\\
Joint R@10 & 98.43 & 98.19 & 92.31 & 92.43\\
Joint R@20 & 98.84 & 98.79 & 95.58 & 95.43\\
Joint R@50 & 99.44 & 99.35 & 97.70 & 97.77\\
Joint MRR & 92.07 & 91.51 & 73.39 & 73.50\\
Joint nDCG@10 & 0.9362 & 0.9314 & 0.7776 & 0.7788\\
Normalized rank $\downarrow$ & 0.001660 & 0.001910 & 0.008991 & 0.008880\\
Drug R@1 & 89.83 & 88.96 & 66.72 & 66.83\\
Drug R@5 & 97.98 & 97.78 & 89.61 & 89.77\\
Drug R@10 & 98.94 & 98.77 & 94.10 & 94.20\\
Drug MRR & 93.57 & 92.97 & 76.61 & 76.69\\
Dose Acc. & 96.47 & 96.44 & 91.47 & 91.51\\
Log-dose error $\downarrow$ & 0.0495 & 0.0508 & 0.1120 & 0.1110\\
\bottomrule
\end{tabular*}
\end{table}

\end{document}